\documentclass{article}
\usepackage{arxiv}
\usepackage[utf8]{inputenc}
\usepackage[T1]{fontenc}
\usepackage{hyperref}
\usepackage{url}
\usepackage{booktabs}
\usepackage{amsfonts}
\usepackage{nicefrac}
\usepackage{microtype}      
\usepackage{lipsum}
\usepackage{graphicx}
\usepackage[numbers,sort&compress]{natbib}
\usepackage{doi}
\usepackage{float}
\usepackage{makecell}
\usepackage[normalem]{ulem}
\usepackage{cancel}
\usepackage{capt-of}
\usepackage{longtable}
\usepackage{caption}
\usepackage{enumitem}
\usepackage{amsmath} 

\usepackage{soul}
\usepackage{xcolor}

\definecolor{aigreen}{RGB}{210,244,211} 
\definecolor{aired}{RGB}{255,180,181} 
\definecolor{lightred}{rgb}{1,0.9,0.9}
\definecolor{lightgrayhl}{RGB}{220,220,220}

\sethlcolor{aigreen}
\newcommand{\hlred}[1]{{%
  \sethlcolor{lightred}%
  \hl{#1}%
}}
\newcommand{\hlgray}[1]{{%
  \sethlcolor{lightgrayhl}%
  \hl{#1}%
}}

\usepackage{mdframed}

\title{GraphAgents: Knowledge Graph–Guided Agentic AI for Cross-Domain Materials Design}

\date{} 					

\author{
\href{https://orcid.org/0009-0003-9266-4137}{\includegraphics[scale=0.06]{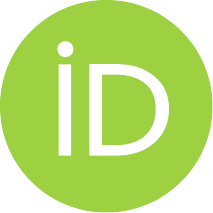}\hspace{1mm}Isabella A. Stewart} \\
Department of Civil and Environmental Engineering \\
Massachusetts Institute of Technology \\
Cambridge, MA, USA
\And
\href{https://orcid.org/0009-0005-6969-1852}{\includegraphics[scale=0.06]{orcid.pdf}\hspace{1mm}Tarjei Paule Hage} \\
Department of Mechanical Engineering \\
Massachusetts Institute of Technology \\
Cambridge, MA, USA
\And
\href{https://orcid.org/0000-0001-5587-4547}{\includegraphics[scale=0.06]{orcid.pdf}\hspace{1mm}Yu-Chuan Hsu} \\
Department of Civil and Environmental Engineering \\
Massachusetts Institute of Technology \\
Cambridge, MA, USA
\And
\href{https://orcid.org/0000-0002-4173-9659}{\includegraphics[scale=0.06]{orcid.pdf}\hspace{1mm}Markus J. Buehler} \\
Department of Civil and Environmental Engineering \\
Department of Mechanical Engineering \\
Schwarzman College of Computing \\
Massachusetts Institute of Technology \\
Cambridge, MA, USA \\
\texttt{mbuehler@mit.edu}
}

\makeatletter
\providecommand{\headeright}{}
\providecommand{\undertitle}{}
\providecommand{\shorttitle}{}
\makeatother

\hypersetup{
pdftitle={GraphAgents: Knowledge Graph–Guided Agentic AI for Cross-Domain Materials Design},
pdfsubject={q-bio.NC, q-bio.QM},
pdfauthor={M.J. Buehler},
pdfkeywords={Large Language Models, Multi-Agent AI, Knowledge Graphs, Hypothesis Generation, PFAS, Sustainable Materials Design}
}

\begin{document}
\maketitle
\begin{abstract}
Large Language Models (LLMs) promise to accelerate discovery by reasoning across the expanding scientific landscape. Yet, the challenge is no longer access to information but connecting it in meaningful, domain-spanning ways. In materials science, where innovation demands integrating concepts from molecular chemistry to mechanical performance, this is especially acute. Neither humans nor single-agent LLMs can fully contend with this torrent of information, with the latter often prone to hallucinations. To address this bottleneck, we introduce a multi-agent framework guided by large-scale knowledge graphs to find sustainable substitutes for per- and polyfluoroalkyl substances (PFAS)—chemicals currently under intense regulatory scrutiny. Agents in the framework specialize in problem decomposition, evidence retrieval, design parameter extraction, and graph traversal, uncovering latent connections across distinct knowledge pockets to support hypothesis generation. Ablation studies show that the full multi-agent pipeline outperforms single-shot prompting, underscoring the value of distributed specialization and relational reasoning. We demonstrate that by tailoring graph traversal strategies, the system alternates between exploitative searches focusing on domain-critical outcomes and exploratory searches surfacing emergent cross-connections. Illustrated through the exemplar of biomedical tubing, the framework generates sustainable PFAS-free alternatives that balance tribological performance, thermal stability, chemical resistance, and biocompatibility. This work establishes a framework combining knowledge graphs with multi-agent reasoning to expand the materials design space, showcasing several initial design candidates to demonstrate the approach.
\end{abstract}


\section{Introduction}
\label{sec:introduction}
A central challenge in developing alternative materials is the narrow domain specialization of human scientists. Polymer scientists, for instance, are highly trained to optimize polymers within well-defined design frameworks, yet this very depth often comes at the cost of breadth. Without a cross-disciplinary perspective that bridges polymers with other materials, such as ceramics, metals, biomaterials, or emerging hybrid systems, experts may struggle to envision replacements that draw on principles or insights outside their immediate field. 

This siloed structure of human knowledge limits scientific discovery in two key ways. It narrows the ability to scan the full landscape of materials science and engineering for viable substitutes, and it reduces the chance encounters between disciplines where novel ideas often emerge. As a result, even when the demand for sustainable alternatives is urgent, the process of hypothesis generation remains slowed by the inherent bounds of human expertise and imagination.

The challenge is particularly visible in efforts to identify substitutes for materials that are difficult to replace. Among these are per- and polyfluoroalkyl substances, commonly referred to as PFAS, a class of synthetic chemicals valued for their durability, chemical resistance, and water- and oil-repellent properties \cite{shabanian_durable_2023, leung_physicochemical_2023}. These same characteristics contribute to the environmental persistence of certain PFAS compounds and have prompted increasing regulatory attention in parts of the world \cite{zheng_per-_2021, henderson_analysis_2024, cousins_high_2020}. Despite this evolving landscape, PFAS remain widely used across industries, as few alternative materials offer a comparable combination of properties.

The growing performance of large language models (LLMs) \cite{Vaswani2017attention, touvron_llama_2023, wei_emergent_2022, zhao_survey_2023, teubner_welcome_2023, naveed_comprehensive_2023} presents a compelling strategy to reason over the massive corpus of scientific knowledge \cite{zhang_exploring_2025, rueda_understanding_2025} in ways that extend beyond the limitations of human-driven analysis. In particular, LLM-driven multi-agent frameworks enable the orchestration of domain-specialized agents to interact and coordinate toward reaching a common goal. These systems are poised for the requirements of materials design problems, which fundamentally demand a multi-step workflow wherein agents work together like a group of collaborative scientists. Unlike traditional machine learning techniques where a single AI model trains once on relevant data, these multi-agent systems can adapt to incoming data and solicit new data on the fly to perform complex tasks. X-LoRA demonstrates how LLMs can be continually adapted in a modular, mixture-of-experts manner by routing each query through specialized low-rank adapter ``experts'', enabling efficient, on-the-fly domain specialization (e.g., for protein mechanics and molecular design) without retraining the full base model~\cite{Buehler2024XLoRA}. Agents can also be integrated with external tools such as physics-based simulators to impose physics constraints, as well as massive databases of scientific data or knowledge bases. This approach has already proven successful in advancing scientific discovery, with applications spanning mechanical engineering \cite{ni_mechagents_2024, Buehler2023MechGPTModalities}, protein design \cite{ghafarollahi_protagents_2024, ghafarollahi2025sparks}, and materials design in areas such as bioinspired and inorganic systems \cite{luu_bioinspiredllm_2024,ghafarollahi2025sparks, luu_generative_2025, stewart_molecular_2025, stewart_higher-order_2026}.

Given these earlier results, there is growing interest in developing agent architectures that coordinate domain-specialist agents to address complex scientific challenges. Research directions such as group-chat structures, dynamic agent access, chat-memory, and adaptive task and tool allocation represent promising avenues for advancement.

An immediate priority is finding ways to extend agent access and use of large-scale knowledge sources. Open research opportunities remain for developing more efficient representations of knowledge in generative AI for science, as originally suggested in \cite{buehler_generative_2024,buehler_accelerating_2024}. A notable strategy is the incorporation of ontological knowledge graphs \cite{marquis_category_2023, mac1998categories, giesa_category_2012, markovic_optimizing_2025, sequeda_knowledge_2025,schilling-wilhelmi_text_2024,Buehler2025GraphPRefLexOR}, a structure in which concepts are represented as nodes and their relations as edges. Recent applications have demonstrated how LLMs can distill vast amounts of scientific text into these structured maps \cite{buehler_accelerating_2024,Buehler2025Chaos}. Graph  representations have also been utilized to develop graph-native reasoning models, as suggested in~\cite{Buehler2025GraphPRefLexOR}. Within agentic frameworks, such graphs become callable tools for in-context learning, enabling agents to retrieve knowledge dynamically and engage with massive scientific data \cite{ghafarollahi2025sciagents, stewart_higher-order_2026}. Because this approach does not rely on training, knowledge is maintained independently of model weights, ensuring transparency, facilitating updates, mitigating drift during inference, and reducing computational overhead. Knowledge graphs also allow retrieval-augmented generation (RAG) to be performed more efficiently than with unstructured text. Condensing large corpora into structured representations enables progress toward a scalable world model that integrates knowledge across domains. Further, relationships are represented explicitly in these graphs, making it possible to leverage the graph structure creatively to both uncover previously unknown connections and gain new insights into the organization of knowledge within a field.

A compelling application of this approach is the use of knowledge graphs as a substrate for idea mining \cite{buehler_generative_2024,buehler_accelerating_2024, ghafarollahi2025sciagents}. Analogous to mapping neurons and their synaptic connections, agents identify nodes within the graph and traverse the connectivities between them to uncover new relationships. This approach demonstrates the synergy of ontological knowledge graphs with the generative and reasoning capabilities of LLMs for idea generation. However, while single agents can contribute creative initial ideas, not every idea constitutes a robust hypothesis. 
Existing KG-based idea-mining approaches can surface novel associations, but they do not, by themselves, provide a reliable mechanism for (i) decomposing a design problem into verifiable sub-questions, (ii) grounding each sub-question in traceable evidence, and (iii) integrating cross-subtask constraints into a single testable hypothesis. This motivates an agentic architecture in which knowledge graphs are not only retrieval indices but controllable reasoning substrates.
To address this need, we introduce GraphAgents, a multi-agent framework that couples evidence-grounded retrieval with controllable graph traversal to support structured, traceable hypothesis generation and refinement for materials design.

\subsection{Model Framework}

\begin{figure}[H]
    \centering
    \includegraphics[width=1\linewidth]{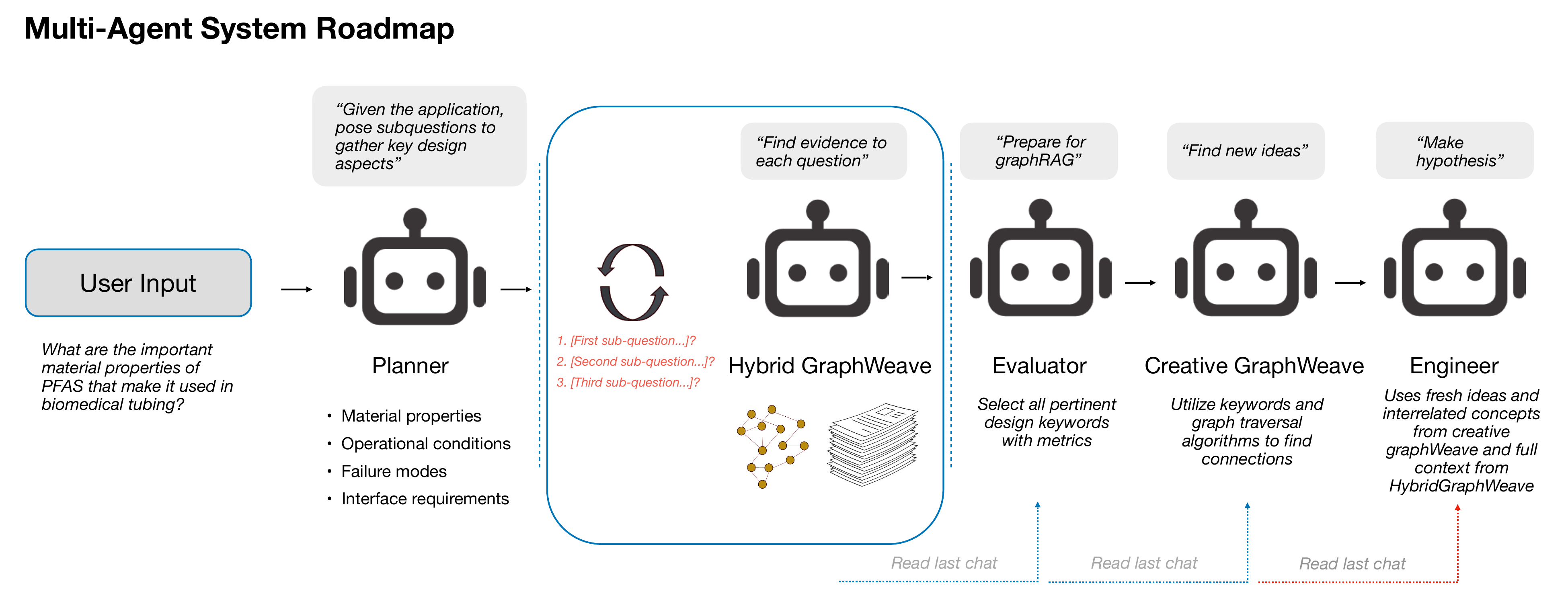}
    \caption{Overview of the proposed agentic pipeline for materials discovery. The system begins with user input, which the Planner decomposes into sub-questions focusing on design parameters like material properties, operational conditions, failure modes, and interface requirements. The Hybrid GraphWeave agent retrieves evidence to answer each sub-question, from which the Evaluator selects pertinent design keywords with associated metrics. The Creative GraphWeave agent explores graph traversal pathways using the Evaluator keywords as stopping points to uncover novel connections and ideas. These are then synthesized by an Engineer into hypotheses that integrate both creative insights and contextual evidence. Bottom arrows illustrate how agents selectively retrieve and incorporate earlier responses as context for their own reasoning.}
    \captionsetup{skip=5pt}
    \label{fig:pipeline}
\end{figure}

\begin{figure}[htbp]
    \centering
    \includegraphics[width=1\linewidth]{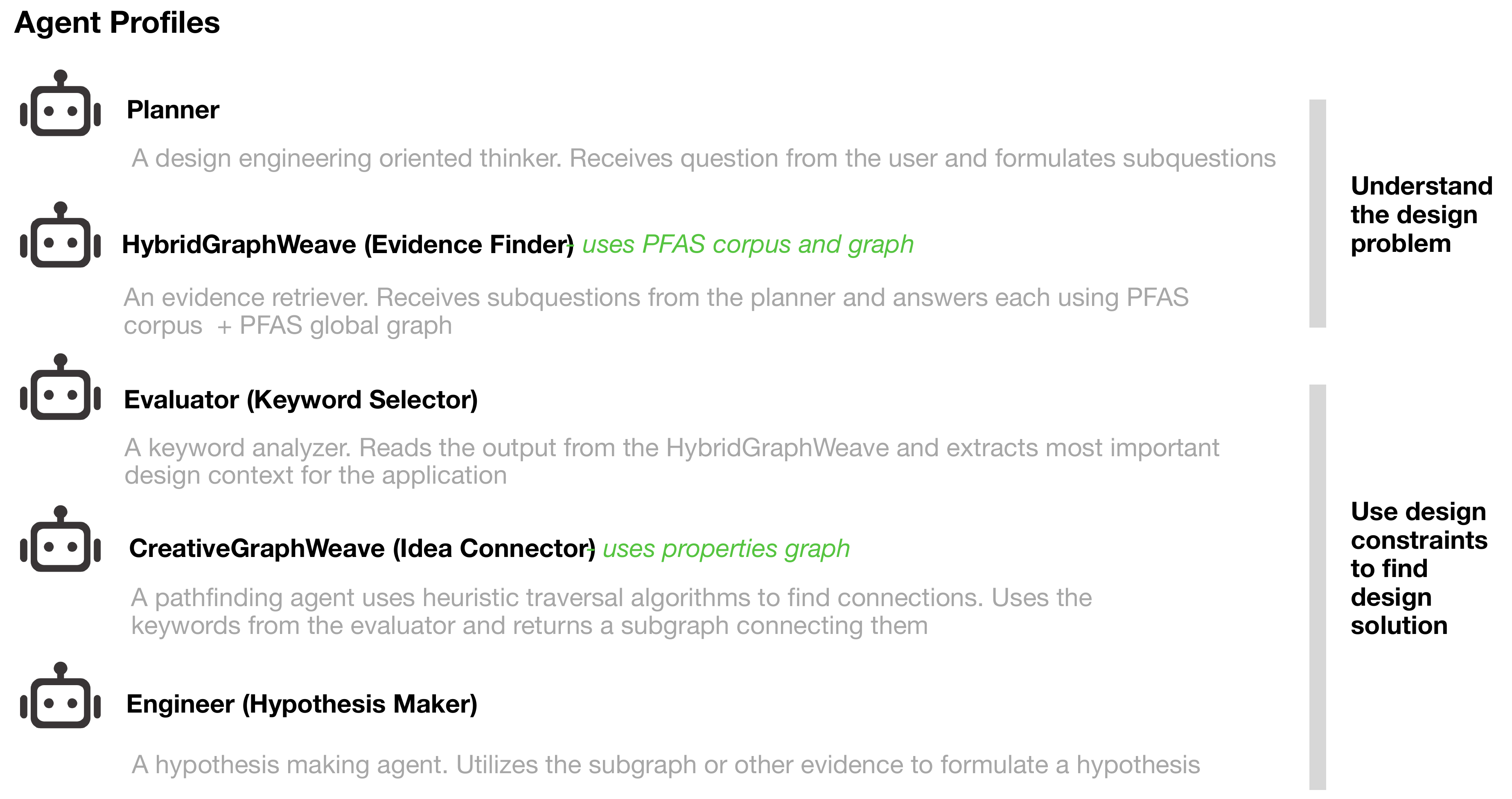}
    \caption{Summary of the roles of each agent in the multi-agent system, from problem decomposition (Planner, HybridGraphWeave) to keyword extraction (Evaluator), idea generation (CreativeGraphWeave), and hypothesis formulation (Engineer).}
    \label{fig:agent_profiles}
\end{figure}

Our framework integrates knowledge graph construction with a multi-agent system to support systematic hypothesis generation. Large corpora of scientific literature are distilled into two complementary knowledge graphs: a PFAS-Specific Knowledge Graph, for depth in the target domain, and a broader Material Properties Knowledge Graph for cross-domain exploration. These structured resources are coupled with a sequence of specialized agents that decompose design problems, retrieve and synthesize evidence from the collected corpus and developed knowledge graphs, extract key design parameters, and traverse graphs with heuristic algorithms to uncover novel relationships. The final stage integrates these insights into hypotheses that are both creative and grounded in prior knowledge. This combination of structured knowledge representation and coordinated agent reasoning provides a scalable foundation for scientific discovery. An overview of the full pipeline can be seen in Figure~\ref{fig:pipeline} with a profile of each agent shown in Figure~\ref{fig:agent_profiles}. 

This systematic approach is designed to emulate deep scientific thinking to accelerate the search for sustainable material alternatives. By combining domain expertise with methods that connect otherwise isolated pockets of knowledge, it becomes possible to broaden the search space, uncover overlooked candidate materials, and generate more creative, testable hypotheses.

\subsection{Outline of the Paper}

The subsequent sections of this paper are organized as follows. Section~\ref{sec:results} presents the results and discussion, including the design of the multi-agent system, knowledge graph construction, and hypothesis generation experiments using heuristic traversal algorithms. This section also reports ablation studies that evaluate the contributions of individual agents to the overall system performance. Section~\ref{sec:conclusion} concludes the paper with a summary of findings and discusses future directions for extending the framework. Finally, Section~\ref{sec:methods} provides details of the materials and methods, including model implementation, knowledge graph generation, and experimental setup.

\section{Results and Discussion}
\label{sec:results}

\subsection{Design of the Agentic System}

Our agentic system follows a sequential framework that begins with design-problem analysis and proceeds to hypothesis generation through deeper reasoning. We begin with an initial user input centered on the PFAS application of biomedical tubing, providing a tractable test case to assess the agentic system. Our approach emphasizes this test case over the pursuit of a comprehensive PFAS substitute, concentrating instead on whether the proposed material can robustly satisfy the application’s critical requirements, which hinge on a vital subset of PFAS properties.

Given that materials discovery is fundamentally a multi-objective, multi-scale optimization problem, the design-problem analysis phase involves first breaking down the larger problem into subproblems, endowing agents with sufficient background knowledge about the required material design parameters. Accordingly, the planner agent is the first to enter the conversation with the user, tasked with analyzing the user’s input and decomposing the overarching question into intermediate subquestions and subtasks.

Figure~\ref{fig:planner-example} illustrates the case study of biomedical tubing, showing both the initial user query and the smaller subquestions generated by the planner agent. The planner is configured to identify as a design engineer, focusing exclusively on questions that clarify critical design parameters such as material properties, operational environments, potential failure modes, and system interfaces. 

\begin{figure}[htbp]
  \centering
  \begin{mdframed}[
    frametitle={User–Planner Interaction},
    frametitlefont=\bfseries\sffamily\small,
    backgroundcolor=white,
    linecolor=black,
    linewidth=0.5pt,
    innerleftmargin=6pt,
    innerrightmargin=6pt,
    innertopmargin=6pt,
    innerbottommargin=6pt
  ]
    \ttfamily\scriptsize  
    \textbf{\hl{User}}: What are the specific material properties of PFAS that make it used for the application of biomedical tubing?
    
    \medskip
    
    \textbf{\hlred{Planner}}: To understand how PFAS enables its application in biomedical tubing, let's break down the inquiry into specific design sub-questions focusing on intrinsic material properties. Here are three simple design sub-questions:
    
    \begin{enumerate}
      \item What is the tensile strength of PFAS that allows it to maintain structural integrity under various pressure conditions within biomedical tubing applications?
      \item How does the low friction coefficient of PFAS contribute to its use in biomedical tubing, particularly in terms of reducing occlusions and improving flow rates?
      \item What is the thermal stability range of PFAS that ensures its dimensional stability and durability when exposed to sterilization processes or varying temperatures during medical procedures?
    \end{enumerate}
    
    \textbf{KEYWORDS}: tensile strength, friction coefficient, thermal stability, biomedical tubing, sterilization resistance, chemical inertness, biocompatibility, non-toxicity, flexibility, kinking resistance.
    
    \textbf{SYNONYMS:}  mechanical properties, surface energy, temperature range, medical devices, implantable devices, perfluorinated compounds, fluoropolymers.
  \end{mdframed}
  \caption{User–Planner Interaction. The Planner agent decomposes a broad user query on PFAS in biomedical tubing into sub-questions that guide the search for more precise answers.}
  \label{fig:planner-example}
\end{figure}

\begin{figure}[htbp]
    \centering
    \includegraphics[width=1\linewidth]{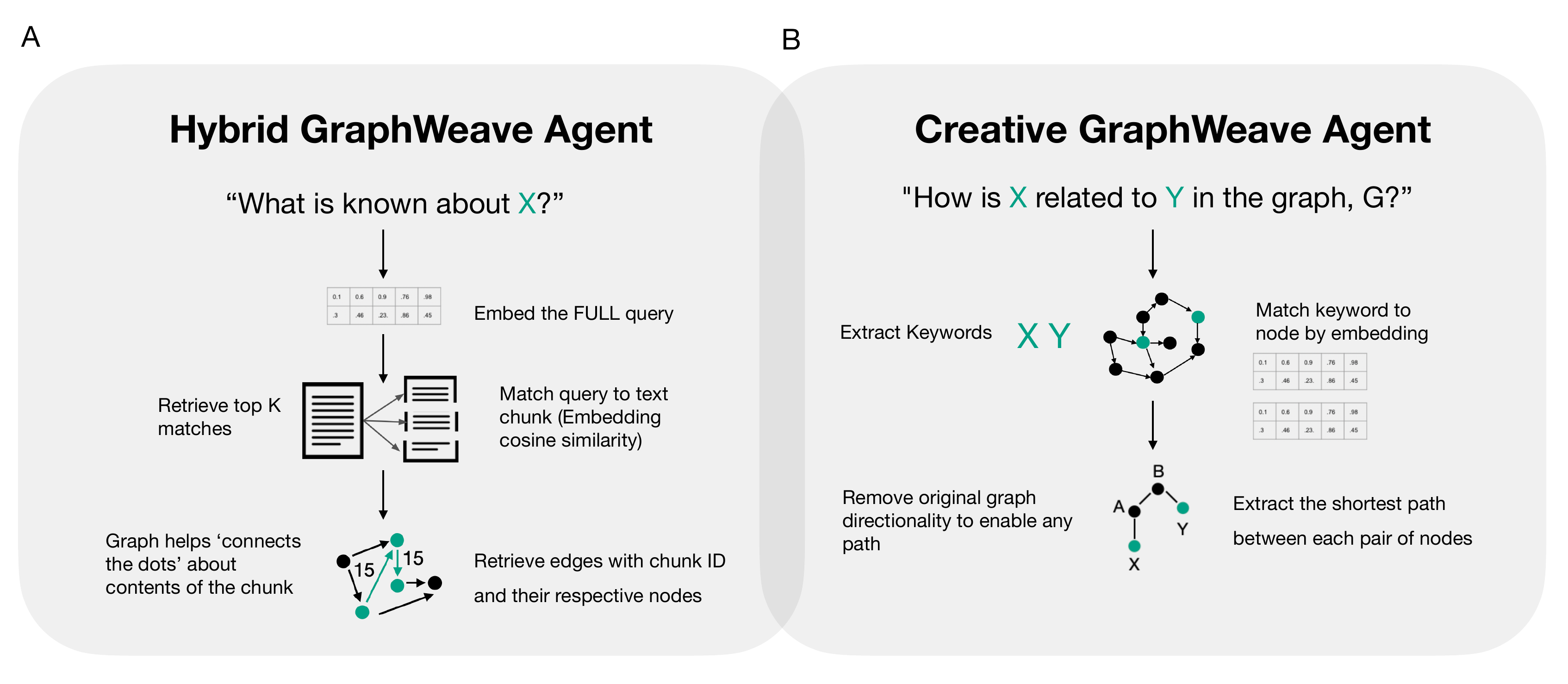}
    \caption{Complementary tool-calling workflows of the (A) Hybrid GraphWeave agent and (B) Creative GraphWeave agent. The Hybrid GraphWeave agent weaves evidence from two sources: the raw PFAS text corpus and the PFAS-Specific Knowledge Graph. It embeds the user query, retrieves the top matching text chunks from the corpus, and links them to corresponding subgraphs in the knowledge graph, thereby providing both textual evidence and relational context. In contrast, the Creative GraphWeave agent focuses on discovering new ideas by exploring and weaving connections between keywords within the knowledge graph. It identifies relevant nodes and applies pathfinding algorithms to assemble subgraphs that highlight potential connections. Together, the two agents provide complementary capabilities: Hybrid GraphWeave ensures evidence-grounded retrieval, while Creative GraphWeave enables exploratory ideation.   
    }
    \label{fig:creative_versus_hybrid}
\end{figure}

Next, we introduce a hybrid approach to knowledge retrieval to answer each of these subquestions using our knowledge base on PFAS. Our Hybrid GraphWeave agent in Figure~\ref{fig:creative_versus_hybrid}(A) utilizes two tools for RAG in order to respond to the planner's questions: the complete raw text corpus on the PFAS knowledge base and the distilled version of the knowledge in the form of the PFAS-Specific Knowledge Graph. The full user input is first embedded into a vector and compared against the embeddings of text chunks from the raw corpus stored in the ChromaDB vector database through cosine similarity. The top k matches (with k = 5) are retrieved and provided in-context as the most relevant information for addressing the subquestion. 

In addition to retrieving raw text, we distill this same content into a knowledge graph. This representation enables agents to more efficiently “connect the dots” by revealing how ideas and concepts are interrelated within the text chunk. Because the PFAS-Specific Knowledge Graph is constructed directly from the PFAS corpus, each edge is tagged with the originating text-chunk ID, ensuring full traceability within the graph’s metadata. This traceability allows us to locate all edges associated with a given ID and gather their corresponding nodes. When these nodes overlap, the collected nodes and edges naturally assemble into subgraphs. Consequently, for each subquestion, the Hybrid GraphWeave agent is provided both the top-k retrieved chunks and the corresponding subgraphs, weaving textual and structural information to guide its responses. The full outputs of the agentic pipeline, detailing the Hybrid GraphWeave agent’s responses to each subquestion, are available in the supplementary files referenced in Code Availability.
 
With the design problem defined, we now turn to cross-domain inspiration by examining adjacent fields of knowledge to identify new materials or processing techniques capable of meeting our required design parameters. To explore the broader web of property relationships across diverse domains, we employ our Material Properties Knowledge Graph built from corpora identified through PFAS property keywords selected by human polymer experts listed in Table~\ref{tab:pfas_keywords}. 

Using the previously gathered background knowledge on the design problem, we first call an Evaluator agent that processes the responses from the Hybrid GraphWeave agent and extracts the most pertinent design keywords and required parameters, along with any detailed metrics stored in the text. An example of this set derived from the evidence gathered by Hybrid GraphWeave agent is shown in Figure ~\ref{fig:evaluator}. Next, our assembled design keywords from the evaluator are embedded into the same vector space as composed for the Material Properties Knowledge Graph and the closest matching nodes in the Material Properties Knowledge Graph are identified and collected. 

The Creative GraphWeave agent in Figure~\ref{fig:creative_versus_hybrid}(B) enumerates paths across every possible permutation of these node pairs, subsequently composing and weaving them into an expanded subgraph. By linking these keywords, we effectively traverse the global knowledge graph, uncovering adjacent fields, broader context, and related concepts that can serve as fertile ground for novel insights.

At this stage, the Engineer agent can reason over these interconnections, exploring how they may be creatively applied to devise sustainable alternatives to PFAS in the specific case of biomedical tubing. In the following section, we examine this agentic interaction in which the Creative GraphWeave agent applies graph traversal algorithms to inspire new hypotheses formulated by the Engineer agent. 

\begin{figure}[htbp]
  \centering
  \begin{mdframed}[
    frametitle={Evaluator Response},
    frametitlefont=\bfseries\sffamily\small,
    backgroundcolor=white,
    linecolor=black,
    linewidth=0.5pt,
    innerleftmargin=6pt,
    innerrightmargin=6pt,
    innertopmargin=6pt,
    innerbottommargin=6pt
  ]
    \ttfamily\scriptsize
    \textbf{\hl{Evaluator}}: 
    tensile strength at 20-30 MPa; friction coefficient in range 0.1 to 0.3; thermal stability in range 250-400°C; dimensional stability at temperatures below 300°C; durability in biomedical tubing applications with exposure to sterilization processes and varying temperatures during medical procedures.
  \end{mdframed}
  \caption{Example of Evaluator agent output. The Evaluator agent selects and formats key material property descriptors with associated ranges or qualitative levels. These extracted keywords and metrics serve as design-relevant features that guide downstream agents in graph traversal and hypothesis generation.}
  \label{fig:evaluator}
\end{figure}

\subsection{Hypothesis Generation with Heuristic Knowledge Graph Traversal Algorithms
}

\begin{figure}[htbp]
  \centering
  \includegraphics[width=0.45\linewidth]{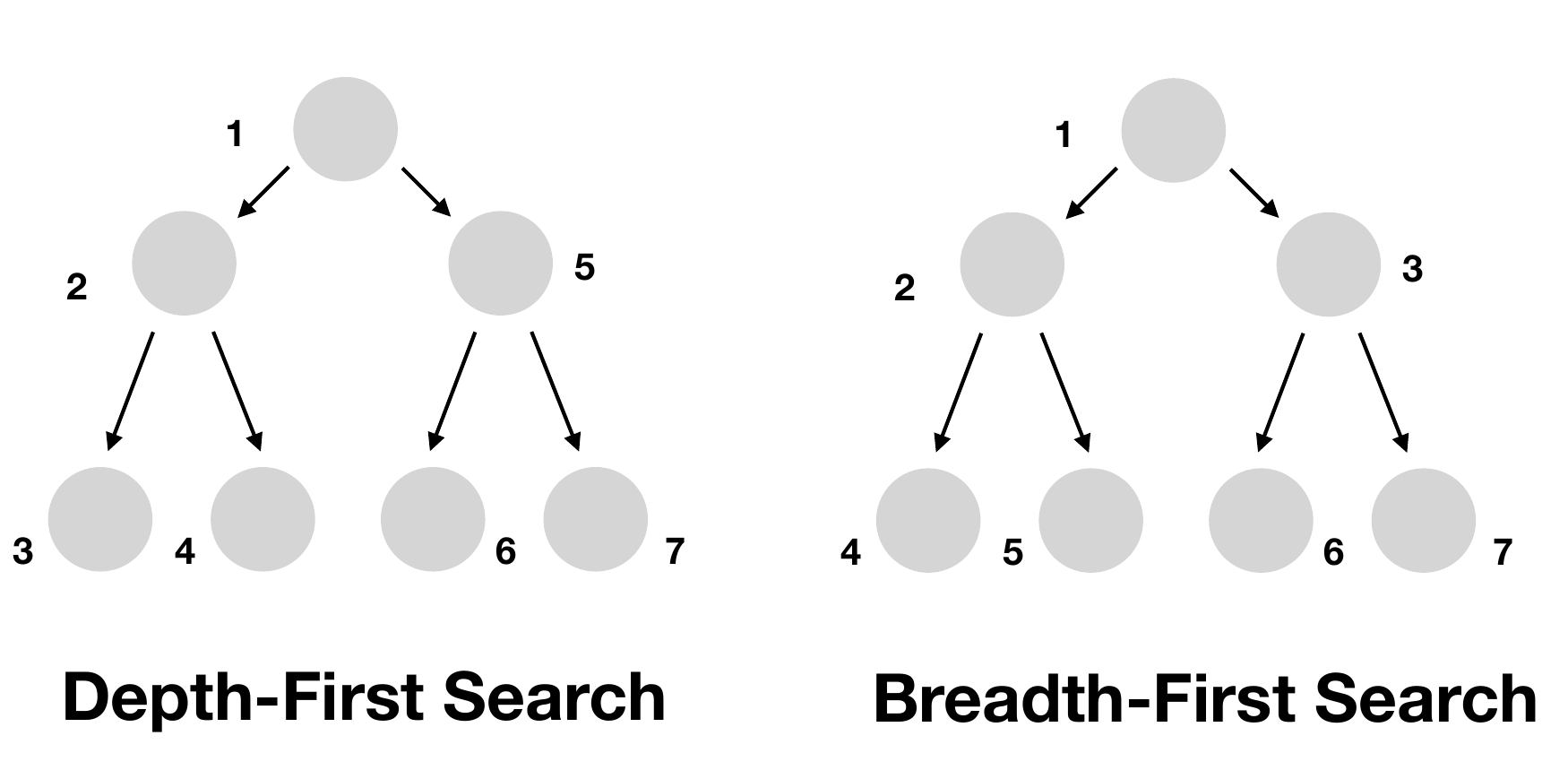}
  \captionsetup{skip=5pt}
  \caption{Comparison of the Depth-First Search (DFS) and Breadth-First Search (BFS) mapping scheme. DFS explores one branch of the graph fully before backtracking, while BFS expands level by level, ensuring the shortest path in an unweighted graph.}
  \label{fig:DFS_BFS_figure}
\end{figure}

In this section, we actively tailor hypothesis generation using four different heuristic graph traversal algorithms, extending the utility of the global Material Properties Knowledge Graph. The choice of traversal algorithm shapes the substrate from which the engineer can draw upon when generating a final hypothesis. By linking nodes across different graph paths, the process can uncover previously unknown connections that map back to the design parameters. The complete pipeline output is provided in the Supporting Information (SI). We build on two approaches to graph-traversal, namely the breadth-first search and depth-first search demonstrated in
Figure~\ref{fig:DFS_BFS_figure}.

Breadth-first search (BFS) guarantees the shortest path in an unweighted graph as it explores nodes in order of increasing distance from the source. Using a queue, BFS first visits all neighbors at distance one, then all nodes at distance two, and so on, expanding outward in layers. This level-by-level exploration ensures that when a node is encountered for the first time, it has been reached through the minimal number of edges possible. As a result, the distances assigned by BFS correspond exactly to the shortest path lengths, forming a shortest-path tree rooted at the source.

Depth-first search (DFS) explores a graph by following one path as far as possible before backtracking. Unlike BFS, which proceeds layer by layer, DFS dives deep into a branch until it reaches a dead end or a previously visited node, then backtracks to explore alternative paths. This strategy is implemented with a stack either explicitly or via recursion, which always expands the most recently discovered node. Unlike BFS, DFS does not guarantee the shortest path between nodes because it may reach a node through a long path before a shorter one is discovered. Instead, DFS is best suited for uncovering structural properties of a graph, such as identifying connected components, maze-solving, classifying edges, or exploring all possible paths.

\begin{figure}[htbp]
    \centering
    \includegraphics[width=1\linewidth]{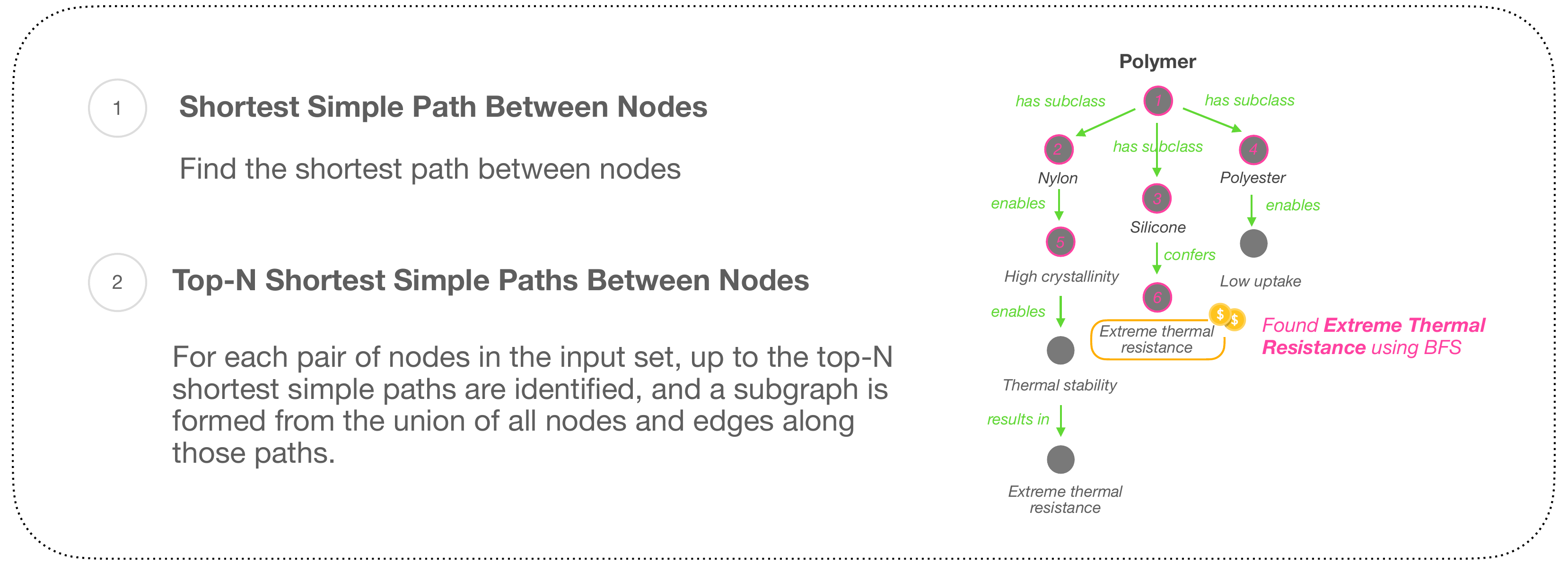}
    \caption{Shortest and Top-N Shortest Simple Paths illustration of how connections between material properties can be mapped into subgraphs from BFS algorithm for candidate discovery. The BFS path is illustrated from the starting node, \textit{polymer}, to the target ending node, \textit{extreme thermal resistance}.}
    \label{fig:shortest_path_figure}
    \captionsetup{skip=5pt}
\end{figure}

A canonical application of BFS is the determination of the shortest simple path, defined as the minimal-length path from a source to a target with no repeated nodes. Figure~\ref{fig:shortest_path_figure} illustrates the Shortest and Top-N Shortest Simple Paths approaches, showing how connections between material properties can be traced through intermediate nodes, where the resulting paths themselves may inspire potential candidate materials. Meanwhile, Figure~\ref{fig:shortest_path} shows how the material keywords, initially extracted by the Evaluator agent, identifies the closest matching nodes in the Material Properties Knowledge Graph, and then computes the shortest path between each node pair using the Shortest Simple Path Algorithm.

\begin{figure*}[t!]
  \begin{mdframed}[
    frametitle={Creative GraphWeave Agent Using Shortest Simple Path Algorithm},
    frametitlefont=\bfseries\sffamily\small,
    backgroundcolor=white,
    linecolor=black,
    linewidth=0.5pt,
    innerleftmargin=6pt,
    innerrightmargin=6pt,
    innertopmargin=6pt,
    innerbottommargin=6pt
  ]
    \ttfamily\scriptsize
    
    \textbf{\hlgray{Extracted material keywords:}}
     ['tensile strength at 20-30 MPa', 'friction coefficient in range 0.1 to 0.3', 'thermal stability in range 250-400°C', 'dimensional stability at temperatures below 300°C', 'durability in biomedical tubing applications with exposure to sterilization processes and varying temperatures during medical procedures']
    
    \textbf{\hlgray{Matched nodes:}}
    ['Tensile strength', 'friction coefficient', 'temperature stability', 'temperature stability', 'Biological durability']
    
    Path between Tensile strength, friction coefficient found as ['Tensile strength', 'adhesion', 'friction coefficient']
    
    Path between Tensile strength, temperature stability found as ['Tensile strength', 'microstructure', 'temperature stability']
    
    Path between Tensile strength, temperature stability found as ['Tensile strength', 'microstructure', 'temperature stability']
    
    Path between Tensile strength, Biological durability found as ['Tensile strength', 'mechanical properties', 'High temperature heat treatment', 'Biological durability']
    
    Path between friction coefficient, temperature stability found as ['friction coefficient', 'temperature stability']
    
    Path between friction coefficient, temperature stability found as ['friction coefficient', 'temperature stability']
    
    Path between friction coefficient, Biological durability found as ['friction coefficient', 'mechanical properties', 'High temperature heat treatment', 'Biological durability']
    
    Path between temperature stability, temperature stability found as ['temperature stability']
    
    Path between temperature stability, Biological durability found as ['temperature stability', 'Perovskite solar cells (PSC)', 'Manufacturing process', 'High temperature heat treatment', 'Biological durability']
    
    Path between temperature stability, Biological durability found as ['temperature stability', 'Perovskite solar cells (PSC)', 'Manufacturing process', 'High temperature heat treatment', 'Biological durability']
    
    Path found ratio = 1.0
    
    ---
    
    \textbf{\hlgray{Please consider these knowledge graph relationships:}}
    
    Tensile strength measures mechanical properties. Tensile strength implies good adhesion. mechanical properties includes Tensile strength. mechanical properties is evaluated by adhesion. mechanical properties includes friction coefficient. mechanical properties correlated to microstructure. High temperature heat treatment improves Biological durability. High temperature heat treatment reduces mechanical properties. temperature stability affects during Fade Test friction coefficient. microstructure affects temperature stability. microstructure affects mechanical properties. microstructure affects friction coefficient. microstructure affects Tensile strength. adhesion affects mechanical properties. adhesion affects friction coefficient. Perovskite solar cells (PSC) lacksProperty temperature stability. Perovskite solar cells (PSC) requires Manufacturing process. friction coefficient changes to friction coefficient. friction coefficient has a positive relationship with adhesion. friction coefficient influenced by mechanical properties. Manufacturing process involves High temperature heat treatment. 
    
  \end{mdframed}
  \caption{Creative GraphWeave agent output with Shortest Simple Path Algorithm. The Creative GraphWeave agent takes material property keywords extracted by the Evaluator and maps them to nodes within the properties knowledge graph.}
  \label{fig:shortest_path}
\end{figure*}

The Material Properties Knowledge Graph does not always yield an exact match for the extracted material keywords. In theory, the closer the retrieved concept aligns with the intended property and its associated metrics, the more effectively it addresses the specific design objective of matching PFAS expected properties. In practice, however, we observe that the matching exhibits both successful semantic alignment and informative mismatches. The keywords "tensile strength at 20-30 MPa" and "friction coefficient in range 0.1 to 0.3" in Figure~\ref{fig:shortest_path} successfully map to their corresponding nodes "Tensile strength" and "friction coefficient," preserving the core concepts while sacrificing numerical  measurement consistency and detail relative to stringent design specifications. 

Other near-fit matches introduce some shifts in semantic emphasis and physical interpretation. The keyword "thermal stability in range 250–400°C" in Figure~\ref{fig:shortest_path} maps to "temperature stability," introducing a subtle semantic divergence between thermal behavior, defined as a material’s response to heat, and temperature stability, understood as its resistance to temperature fluctuations. Similarly, "dimensional stability at temperatures below 300°C" also maps to "temperature stability," discarding the dimensional aspect of maintaining physical shape and size under thermal stress, not merely thermal endurance.

Most notably, the verbose key node "durability in biomedical tubing applications with exposure to sterilization processes and varying temperatures during medical procedures" in Figure~\ref{fig:shortest_path} collapses to "Biological durability." While this captures the biological context, the mapping strips away application-specific constraints such as sterilization compatibility and cyclic temperature exposure. This blurred match may encourage the system to explore biological durability broadly rather than focusing on the precise environmental stresses inherent to biomedical tubing applications. 

By incorporating these exact and near-fit matches, the system embraces both precision (when available) and exploratory breadth (when not), striking a balance between structured reasoning and serendipitous discovery that may yield unexpected insights. 

Next, the Engineer is tasked with leveraging the extracted knowledge graph relationships to formulate a hypothesis for a PFAS-free material or composite. Its output is constrained to a standardized format that includes: (i) a proposed material or composite hypothesis, (ii) a justification explicitly mapped to design requirements and associated metrics, (iii) expected material properties spanning mechanical, thermal, chemical, transport, and biological domains, (iv) foreseeable implementation challenges such as cost, scalability, or manufacturability, and (v) explicit references to the knowledge graph paths that underpin the reasoning chain. Figure~\ref{fig:engineer_shortest_path} illustrates the full hypothesis output obtained by following the shortest simple path identified earlier by the Creative GraphWeave agent.

\begin{figure*}[t!]
  \begin{mdframed}[
    frametitle={Generated Hypothesis as a Result of Shortest Simple Path Algorithm},
    frametitlefont=\bfseries\sffamily\small,
    backgroundcolor=white,
    linecolor=black,
    linewidth=0.5pt,
    innerleftmargin=6pt,
    innerrightmargin=6pt,
    innertopmargin=6pt,
    innerbottommargin=6pt
  ]
    \ttfamily\scriptsize
    
    \textbf{\hlgray{Hypothesis:}}
    
    A novel, PFAS-free composite material is proposed for biomedical tubing applications, consisting of a poly(lactic acid) (PLA) matrix reinforced with cellulose nanofibers (CNFs) and functionalized with a thin layer of polydopamine (PDA). This formulation integrates components into a stable system by leveraging the biocompatibility and flexibility of PLA, the mechanical reinforcement provided by CNFs, and the surface modification capabilities of PDA. The expected performance metrics include enhanced tensile strength ($>$50~MPa), improved thermal stability ($T_g > 60$\textdegree C), reduced gas permeability ($<10^{-13}$~cm$^3\cdot$cm/cm$^2\cdot$s$\cdot$Pa), and excellent chemical resistance to various solvents and biological fluids.
    
    \textbf{\hlgray{Justification:}}
    
    1. Mechanical Performance: The incorporation of CNFs into the PLA matrix is expected to enhance the tensile strength and modulus of the composite, satisfying the requirement for mechanical integrity in biomedical tubing applications.
    
    2. Thermal Stability: High-temperature heat treatment can be applied to improve the biological durability of the composite while maintaining its mechanical properties, as suggested by the knowledge graph path "High temperature heat treatment improves Biological durability.
    
    3. Surface Properties: The PDA coating is expected to reduce the friction coefficient and enhance the wettability of the composite surface, which is crucial for biomedical applications where biocompatibility and non-fouling properties are essential.
    
    \textbf{\hlgray{Expected Material Properties for Experimental Evaluation:}}
    
    - Mechanical performance:  Tensile strength = 55 MPa, Modulus = 3.5 GPa
    
    - Thermal behavior: Tg = 65°C, Melting point = 180°C
    
    - Transport properties: Gas permeability = $5 \times 10^{-14}$~cm$^3\cdot$cm/cm$^2\cdot$s$\cdot$Pa, OTR = 100~cm$^3$/m$^2\cdot$day
    
    - Chemical stability/resistance: Resistant to pH 1-12 solutions, various organic solvents, and biological fluids
    
    - Surface/interfacial properties: Friction coefficient = 0.2, Contact angle = 60°
    
    - Biological compatibility: Non-cytotoxic, hemocompatible, and biodegradable
    
    \textbf{\hlgray{Foreseeable Implementation Challenges:}}
    
    1. Scalability: Large-scale production of CNFs and uniform dispersion within the PLA matrix may be challenging. 
    
    2. Cost: The use of PDA as a surface modifier may increase the overall cost of the material. 
    
    3. Regulatory Barriers: Biocompatibility and biodegradability tests may be required to ensure regulatory compliance.
    
    \textbf{\hlgray{Knowledge Graph Reasoning Path(s) Used:}}
    
    The knowledge graph paths used in this hypothesis include "mechanical properties → Tensile strength," "High temperature heat treatment improves Biological durability," and "microstructure affects mechanical properties." The relationships between these nodes led to the selection of a PLA-CNF-PDA composite as a potential replacement for PFAS-based materials in biomedical tubing applications. Specifically, the path "adhesion affects mechanical properties" suggested the use of PDA as a surface modifier to enhance adhesion and reduce friction coefficient, while the path "microstructure affects temperature stability" informed the selection of CNFs as a reinforcement material to improve thermal stability.
    
  \end{mdframed}
  \caption{Engineer agent output using the Shortest Simple Path Algorithm.}
  \label{fig:engineer_shortest_path}
\end{figure*}

The proposed solution is an explicitly multi-objective composite material consisting of a poly(lactic acid) (PLA) matrix reinforced with cellulose nanofibers (CNFs) and functionalized with a thin polydopamine (PDA) layer. The hypothesis strategically leverages several clusters of relationships from the knowledge graph to construct a PFAS-free composite system. PLA is selected as the base matrix for its inherent biocompatibility, flexibility, and biodegradability, providing a foundation suitable for biomedical applications. CNFs contribute synergistically by enhancing tensile strength and modulus while improving thermal stability through microstructural reinforcement, as supported by the knowledge graph path linking microstructure to both mechanical properties and temperature stability. PDA is incorporated as a surface modifier to reduce friction coefficient and enhance wettability, leveraging the graph connection between adhesion and mechanical properties to improve interfacial performance. Beyond these individual contributions, the proposal draws on relational edges such as the links between high-temperature heat treatment and biological durability, showing how thermal processing can enhance longevity without compromising mechanical integrity. Together, these paths justify a stable, multifunctional system that balances mechanical strength, thermal robustness, surface properties, and biological compatibility as a viable PFAS alternative for biomedical tubing applications. 

Applying this traversal method to our knowledge graph, we observed that the shortest paths, being considerably short, did not always provide enough contextual material for the Engineer to formulate a well-informed hypothesis. For instance, as can be seen in Figure~\ref{fig:shortest_path} we encounter self-referential paths such as “friction coefficient changes to friction coefficient” or statements made out of context like “adhesion affects mechanical properties.” These hollow edges can dominate the shortest paths, limiting the Engineer’s reasoning. To address this limitation, we extend the approach to the Top-N Shortest Simple Paths (with a default of N=5) in Figure~\ref{fig:engineer_topN}, enabling us to retrieve multiple distinct paths that reveal broader connectivity between node pairs. 

Compared to the shortest-path hypothesis, which centered on a PLA-CNF-PDA composite and emphasized biocompatibility and surface modification through minimal connections, the Top-N Shortest Simple Paths ($N=5$) yielded a more multidimensional proposal, integrating additional material components and exhibiting richer cross-functional synergies. By integrating thermoplastic polyurethane (TPU), cellulose nanocrystals (CNCs), and alumina (Al$_2$O$_3$) nanoparticles, the Top-N hypothesis captured a broader set of property domains including mechanical flexibility, thermal stability, protein resistance, and surface biocompatibility. TPU is selected as the base matrix for its superior flexibility and mechanical strength compared to PLA, providing a more robust foundation for tubing applications that require both durability and pliability. CNCs contribute by enhancing tensile strength and thermal stability, drawing on multiple knowledge graph paths that connect cellulose-based reinforcement to both mechanical performance and temperature resistance. The Al$_2$O$_3$ nanoparticle coating is incorporated specifically to address protein adsorption and surface biocompatibility, leveraging graph connections between surface modification and biological interactions that were not captured in the single shortest path. Beyond these individual contributions, the Top-N approach reveals deeper relational edges connecting interfacial adhesion, protein resistance, and biocompatibility, demonstrating how expanding beyond the single shortest path allows the agent to leverage diverse clusters of graph relationships. The Top-N strategy results in a more experimentally grounded PFAS-free material hypothesis that balances a wider array of design requirements.

\begin{figure*}[htbp]
  \begin{mdframed}[
    frametitle={Generated Hypothesis as a Result of Top-N Shortest Simple Path Algorithm (N=5)},
    frametitlefont=\bfseries\sffamily\small,
    backgroundcolor=white,
    linecolor=black,
    linewidth=0.5pt,
    innerleftmargin=6pt,
    innerrightmargin=6pt,
    innertopmargin=6pt,
    innerbottommargin=6pt
  ]
    \ttfamily\scriptsize
    
    \textbf{\hlgray{Hypothesis:}}
    
    A novel, PFAS-free composite material is proposed, consisting of a thermoplastic polyurethane (TPU) matrix reinforced with cellulose nanocrystals (CNCs) and modified with a thin layer of alumina (Al2O3) nanoparticles on its surface. This formulation integrates the components into a stable system by leveraging the biocompatibility, flexibility, and mechanical strength of TPU, the reinforcement and thermal stability provided by CNCs, and the improved surface properties and protein resistance offered by the Al2O3 nanoparticle coating. The expected performance metrics include enhanced tensile strength (> 50 MPa), improved temperature stability (up to 250°C), reduced friction coefficient (< 0.1), and increased biocompatibility due to the non-toxic and natural components used.
    
    \textbf{\hlgray{Justification:}}
    
    1. Mechanical Performance: The TPU matrix provides a flexible and mechanically strong base, while the CNCs act as reinforcement, enhancing the tensile strength and stiffness of the composite.
    
    2. Thermal Stability: The incorporation of CNCs and the Al2O3 nanoparticle coating improves the thermal stability of the material, allowing it to withstand high temperatures without significant degradation.
    
    3. Surface Properties: The Al2O3 nanoparticle coating reduces protein adsorption and improves the biocompatibility of the surface, which is crucial for biomedical applications.
    
    \textbf{\hlgray{Expected Material Properties for Experimental Evaluation:}}
    
    - Mechanical performance: Tensile strength = 55 MPa, Modulus = 1.2 GPa
    
    - Thermal behavior: Tg = 180°C, Melting point = 220°C
    
    - Transport properties: Gas permeability = $10^{-13}$~m$^3$/m/s/Pa, OTR = 0.05~cm$^3$/m$^2$/day
    
    - Chemical stability/resistance: Resistant to water, saline solutions, and common biomedical solvents
    
    - Surface/interfacial properties: Friction coefficient = 0.08, Contact angle = 60°
    
    - Biological compatibility: Non-cytotoxic, hemocompatible, and biodegradable
    
    \textbf{\hlgray{Foreseeable Implementation Challenges:}}
    
    1. Scalability: Scaling up the production of CNCs while maintaining their uniformity and quality.
    
    2. Interfacial Adhesion: Ensuring strong interfacial adhesion between the TPU matrix, CNCs, and Al2O3 nanoparticles to prevent delamination under stress or environmental exposure.

    3. Cost: The cost of producing high-quality CNCs and applying the Al2O3 nanoparticle coating may be higher than traditional materials, affecting the economic viability of the composite.
    
    \textbf{\hlgray{Knowledge Graph Reasoning Path(s) Used:}}
    
    The reasoning path involved connections between TPU (biocompatibility + flexibility), CNCs (reinforcement + thermal stability), Al2O3 nanoparticles (surface modification + protein resistance), and the relationships between surface properties, biocompatibility, and mechanical performance. Specifically, the paths included: TPU → biocompatibility + flexibility; Cellulose nanocrystal → reinforcement synergy; Al2O3 → improved surface properties and protein resistance; Surface properties → biocompatibility and protein adsorption. These connections led to the hypothesis of a composite material that integrates these beneficial properties for biomedical tubing applications.
    
  \end{mdframed}
  \caption{Engineer agent output using the Top-N Shortest Simple Path Algorithm with N=5.}
  \label{fig:engineer_topN}
\end{figure*}

\subsubsection{Depth-First Search (DFS) Path Between Nodes}
\begin{figure}[!t]
    \centering
    \includegraphics[width=1\linewidth]{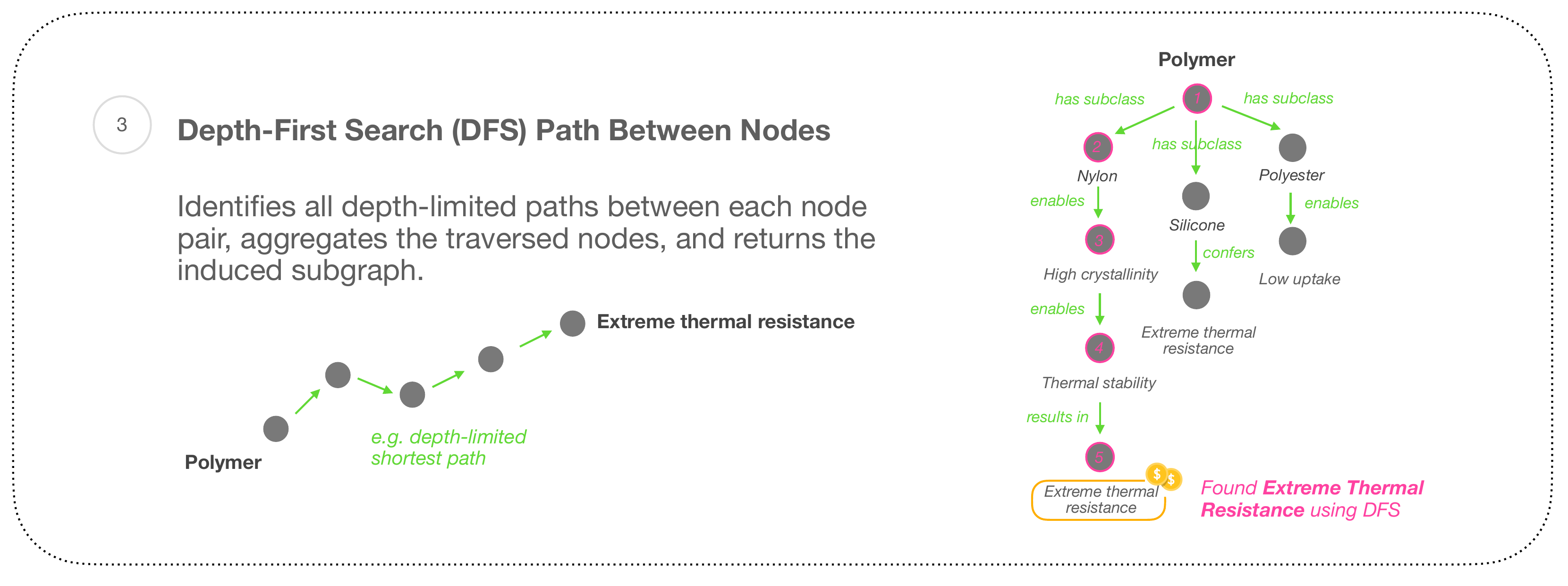}
    \caption{Depth-First Search (DFS) path visualization between material property nodes. DFS explores each branch fully before backtracking, aggregating traversed nodes to form an induced subgraph connecting related properties. The DFS path is illustrated from the starting node, \textit{polymer}, to the target ending node, \textit{extreme thermal resistance}.}
    \label{fig:DFS_figure}
    \captionsetup{skip=5pt}
\end{figure}

In our next experiment, we replace the BFS strategy with a DFS approach applied to our Material Properties Knowledge Graph as shown in Figure~\ref{fig:DFS_figure}. By prioritizing depth before breadth, DFS enriches our hypothesis generation stage by encouraging the accumulation of contextual information along a single path. To avoid the computational expense of traversing to the outer edges of the large knowledge graph, we impose a depth limit of five nodes deep on each branch of the graph, which balances tractability with sufficient depth of exploration. Given the large volume of the output, the Creative GraphWeave Agent’s DFS results are presented in the supplementary.

Figure~\ref{fig:engineer_DFS} shows the full hypothesis output derived from DFS algorithm paths, which proposes a PFAS-free composite material for biomedical tubing applications composed of a natural biopolymer matrix, such as cellulose nanofibers or chitosan, reinforced with multi-walled carbon nanotubes (MWCNTs) and functionalized with TiO\textsubscript{2} nanoparticles. Unlike BFS algorithms that tend to surface more direct associations through flatter paths between node pairs, the DFS approach explores deeper branches of the knowledge graph, resulting in a distinctly hierarchical composite architecture with layered, multidimensional functionality. This depth-prioritized traversal integrates multiple property domains along extended reasoning chains: natural biopolymers provide the biocompatible and flexible structural base with controlled protein adsorption, MWCNTs introduce mechanical reinforcement and electrical conductivity accessed through intermediate reinforcement nodes, and TiO\textsubscript{2} nanoparticles add microstructural stability, corrosion resistance, and additional biocompatibility through deeper functionalization pathways. This depth-first accumulation of context along extended paths enables the identification of less obvious synergies, forming an integrated hypothesis in which mechanical, thermal, electrical, and chemical properties build cumulatively rather than independently.

\begin{figure*}[htbp]
  \begin{mdframed}[
    frametitle={Generated Hypothesis as a Result of Depth-First Search (DFS) Algorithm},
    frametitlefont=\bfseries\sffamily\small,
    backgroundcolor=white,
    linecolor=black,
    linewidth=0.5pt,
    innerleftmargin=6pt,
    innerrightmargin=6pt,
    innertopmargin=6pt,
    innerbottommargin=6pt
  ]
    \ttfamily\scriptsize
    
    \textbf{\hlgray{Hypothesis:}}
    
    A novel, PFAS-free composite material is proposed for biomedical tubing applications, consisting of a blend of natural biopolymers (such as cellulose nanofibers or chitosan) reinforced with Multi-walled carbon nanotubes (MWCNTs) and functionalized with TiO2 nanoparticles. This formulation integrates the components into a stable system by leveraging the biocompatibility, flexibility, and protein adsorption properties of natural biopolymers, the mechanical reinforcement and electrical conductivity provided by MWCNTs, and the improved microstructure, corrosion resistance, and biocompatibility imparted by TiO2. The expected synergistic effects include enhanced tensile strength, temperature stability, and chemical resistance, making this composite a viable alternative to PFAS-based materials in demanding biomedical applications.
    
    \textbf{\hlgray{Justification:}}
    
    1. Mechanical Performance: The incorporation of MWCNTs is expected to significantly enhance the tensile strength and toughness of the composite, potentially exceeding 100 MPa, which is comparable to or better than many PFAS-based materials.
    
    2. Thermal Stability: The presence of TiO2 nanoparticles can improve the thermal stability of the composite by enhancing its microstructure, allowing it to maintain its integrity over a wide temperature range, which is crucial for biomedical applications where sterilization and variable environmental conditions are common.
    
    3. Biocompatibility and Protein Adsorption: Natural biopolymers provide inherent biocompatibility and support cell growth, while TiO2 has been shown to be biocompatible and can interact with proteins in a way that may enhance the composite's overall biological performance.
    
    \textbf{\hlgray{Expected Material Properties for Experimental Evaluation:}}
    
    - Mechanical performance: Tensile strength = 120 MPa, Modulus = 5 GPa
    
    - Thermal behavior: Tg = 180$^{\circ}$C, Thermal stability up to 200$^{\circ}$C
    
    - Transport properties: Gas permeability $< 10^{-13}$~m$^3$/m/s/Pa and WVTR $< 0.1$~g/m$^2$/day.
    
    - Chemical stability/resistance: Resistant to common biomedical solvents and cleaning agents
    
    - Surface/interfacial properties: Friction coefficient = 0.2, Contact angle = 60$^{\circ}$ (indicating moderate hydrophilicity)
    
    - Biological compatibility: Non-cytotoxic, supports cell attachment and growth

    \textbf{\hlgray{Foreseeable Implementation Challenges:}}
    
    1. Scalability and Uniformity: Achieving uniform dispersion of MWCNTs and TiO2 nanoparticles within the natural biopolymer matrix on a large scale could be challenging.
    
    2. Cost and Availability: The cost of high-quality MWCNTs and TiO2 nanoparticles, along with the potential complexity of the fabrication process, might limit the economic viability of this composite for some applications. 
    
    3. Regulatory Approval: Biomedical applications require rigorous testing and regulatory approval, which can be time-consuming and costly.
    
    \textbf{\hlgray{Knowledge Graph Reasoning Path(s) Used:}}
    
    The reasoning path involved connections between natural biopolymers (e.g., cellulose nanofibers, chitosan) and their properties such as biocompatibility, flexibility, and protein adsorption; MWCNTs and their effects on mechanical performance, electrical conductivity, and friction coefficient; TiO2 nanoparticles and their influence on microstructure, corrosion resistance, thermal stability, and biocompatibility. The path also considered the importance of temperature stability, chemical resistance, and cytotoxicity in biomedical applications, leading to the formulation of a composite that addresses these requirements without the use of PFAS or fluorine-containing materials.
    
  \end{mdframed}
  \caption{Engineer agent output using the Depth-First Search (DFS) algorithm.}
  \label{fig:engineer_DFS}
\end{figure*}

\subsubsection{Breadth-First Search (BFS) with Semantic Stop}

\begin{figure*}[htbp]
    \centering
    \includegraphics[width=1\linewidth]{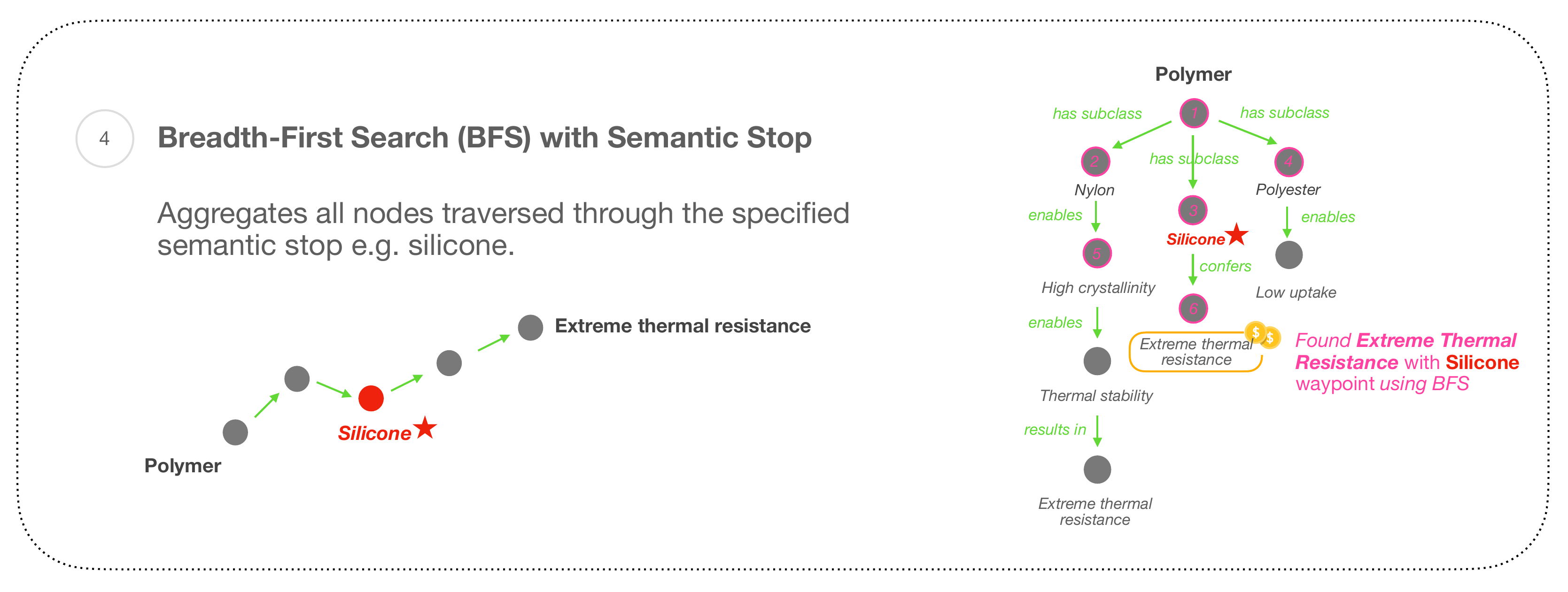}
    \caption{Breadth-First Search (BFS) with a semantic stop shows how paths are directed through the node, aligning property relationships to generate candidate insights. The BFS path is illustrated from the starting node, \textit{polymer}, to the target ending node, \textit{extreme thermal resistance} with semantic stop \textit{Silicone}}
    \label{fig:semantic_stop_figure}
    \captionsetup{skip=5pt}
\end{figure*}

Next, we study an approach to tailoring hypothesis generation by introducing design constraints as semantic stops through which each path is required to pass through. As shown in Figure~\ref{fig:semantic_stop_figure}, the subgraph must include a pre-defined semantic stop. As in the previous algorithms, we begin by collecting the design parameter keywords. For each pair of keyword nodes, we then insert the desired semantic stop as a mandatory waypoint and compute the shortest path between them that traverses this stop. The desired semantic stop is defined by the user. In this test case, the user specifies the design constraint \textit{silk}, which is embedded and matched to the closest node in the graph using cosine similarity of the embeddings. This semantic stop consequently causes all keyword-pair paths to converge through it. All intermediary nodes and edges encountered are subsequently collected as demonstrated in the full output in Figure~\ref{fig:silk_fibroin_paths}.

\begin{figure*}[htbp]
  \begin{mdframed}[
    frametitle={Generated Hypothesis as a Result of Breadth-First Search (BFS) with Semantic Stop 'Silk'},
    frametitlefont=\bfseries\sffamily\small,
    backgroundcolor=white,
    linecolor=black,
    linewidth=0.5pt,
    innerleftmargin=6pt,
    innerrightmargin=6pt,
    innertopmargin=6pt,
    innerbottommargin=6pt
  ]
    \ttfamily\scriptsize
    
    \textbf{\hlgray{Hypothesis:}}
    
    A novel, PFAS-free composite material is proposed for biomedical tubing applications, consisting of a silk fibroin (SF) matrix reinforced with titanium dioxide (TiO2) nanoparticles and integrated into a eutectogel system. This composite leverages the biocompatibility and hydrogen bonding capabilities of SF, the resistance and friction-reducing properties of TiO2, and the temperature stability of eutectogels. The formulation integrates these components through a sol-gel process, where TiO2 nanoparticles are dispersed within a silk fibroin solution, which is then cast into a film or tubing form. The eutectogel component enhances the thermal stability and mechanical strength of the composite. This material system is expected to exhibit improved tensile strength, flexibility, and resistance to degradation under various environmental conditions, making it a viable alternative to PFAS-based materials in biomedical applications.
    
    \textbf{\hlgray{Justification:}}
    
    1. Biocompatibility and Non-toxicity: The use of silk fibroin as the matrix material ensures biocompatibility, which is crucial for biomedical tubing applications. TiO2 nanoparticles are also known for their biocompatibility, further supporting the safety of this composite.
    2. Mechanical Strength and Flexibility: The reinforcement of SF with TiO2 nanoparticles is expected to enhance the tensile strength and modulus of the composite while maintaining its flexibility, which is essential for tubing applications that require both durability and pliability.
    3. Thermal Stability and Resistance: The incorporation of eutectogels into the composite provides enhanced thermal stability, allowing the material to perform consistently across a range of temperatures. TiO2's resistance properties also contribute to the overall stability of the material against environmental stresses.
    
    \textbf{\hlgray{Expected Material Properties for Experimental Evaluation:}}
    
    - \textbf{Mechanical performance:} Tensile strength $\geq$ 50~MPa, modulus $\approx$ 1~GPa, toughness $>$ 10~MJ/m$^3$
    - \textbf{Thermal behavior:} Glass transition temperature ($T_g$) around 200$^\circ$C, thermal stability up to 250$^\circ$C
    - \textbf{Transport properties:} Gas permeability $< 10^{-17}$~m$^3\cdot$m/m$^2\cdot$s$\cdot$Pa, water vapor transmission rate (WVTR) $\approx 10^{-5}$~g/m$^2\cdot$day
    - \textbf{Chemical stability/resistance:} Resistant to pH variations (pH 3--10), stable in aqueous environments and against common biomedical solvents
    - \textbf{Surface/interfacial properties:} Friction coefficient $< 0.1$, contact angle $\approx 60^\circ$ indicating moderate hydrophilicity
    - \textbf{Biological compatibility:} Non-cytotoxic, biodegradable with a degradation rate compatible with the application's requirements
    
    \textbf{\hlgray{Foreseeable Implementation Challenges:}}
    
    1. Scalability and uniformity: Achieving uniform dispersion of TiO2 nanoparticles within the silk fibroin matrix on a large scale could be challenging. 
    2. Cost-Effectiveness: The production cost of this composite, especially considering the purification and processing of silk fibroin and the synthesis of eutectogels, might be higher than traditional materials.
    3. Regulatory Approval: New biomedical materials must undergo rigorous testing and regulatory approval processes, which can be time-consuming and costly.
    
    \textbf{\hlgray{Knowledge Graph Reasoning Path(s) Used:}}
    
    The hypothesis was developed by following the connections between silk fibroin (enabling hydrogen bonding, stabilizing emulsions), TiO2 (reducing friction coefficient, biocompatibility), and eutectogels (exhibiting temperature stability, biocompatibility). The relationships between these components, such as the effect of pH on emulsion stability, hydrogen bonding, and the role of heat treatment in enhancing tensile strength and dimensional stability, were crucial in designing a composite that integrates these materials synergistically to replace PFAS-based systems in biomedical tubing applications.
    
  \end{mdframed}
  \caption{Engineer agent output using the BFS algorithm with a semantic stop. In this case, the semantic stop was set to \textit{Silk}, guiding the breadth-first traversal to prioritize nodes and reasoning chains associated with silk fibroin. The agent generates hypotheses that emphasize silk fibroin's role in material design, particularly its biocompatibility and hydrogen bonding capacity. BFS with semantic stopping enables broader exploration of the knowledge graph while still converging on outputs that satisfy a specific application-driven constraint.}
  \label{fig:engineer_BFS_semantic}
\end{figure*}

The broader utility of fine-tuning the hypothesis with this algorithm lies in its ability to guide search processes toward domain-critical outcomes. By steering BFS with a semantic stop node such as \textit{silk}, the algorithm can prioritize the extraction of hypotheses that emphasize silk fibroin’s role in material design, effectively embedding expert constraints into the reasoning pipeline. This kind of fine-tuning allows the method to operate not just as a blind graph traversal, but as a controllable design tool that aligns its outputs with practical goals, regulatory standards, or performance criteria. Such an approach allows researchers to systematically exploit the vast and complex topology of scientific knowledge while retaining control over which material properties or requirements guide the hypothesis generation process.

\begin{figure}[htbp]
  \begin{mdframed}[
    frametitle={Creative GraphWeave Agent Using Breadth-First Search (BFS) with Semantic Stop at 'Silk'},
    frametitlefont=\bfseries\sffamily\small,
    backgroundcolor=white,
    linecolor=black,
    linewidth=0.5pt,
    innerleftmargin=6pt,
    innerrightmargin=6pt,
    innertopmargin=6pt,
    innerbottommargin=6pt
  ]
    \ttfamily\scriptsize
    
    \textbf{\hlgray{Extracted material keywords:}}
    
    ['chemical resistance at value of high degree', 'low friction coefficient in range of 0.03 to 0.091', 'thermal stability at value of high temperature resistance', 'hydrophobicity at value of low surface energy', 'non-adhesive behavior at value of effortless sliding', 'molecular structure at value of tetrahedral arrangement', 'tribological performance at value of reduced wear and friction', 'surface energy at value of low energy', 'adhesive bonds at value of reduced bonds', 'sample flow at value of smooth flow', 'handling at value of easy handling', 'sterilization methods at value of resistant to sterilization', 'biocompatibility at value of compatible with biological samples']
    
    \textbf{\hlgray{Matched nodes:}}
    
    ['high-temperature resistance', 'friction coefficient', 'high-temperature resistance', 'hydrophobicity', 'adhesion', 'tetragonal structure', 'tribological behavior', 'surface energy', 'covalent bonds', 'Viscosity Measurements', 'efficient packing', 'cryopreservation techniques', 'biocompatibility']
    
    \textbf{\hlgray{Paths involving silk fibroin:}}
    
    Found path from high-temperature resistance to silk fibroin: ['high-temperature resistance', 'Polyimides', 'hydrogen bonding', 'silk fibroin']
    
    Found path from friction coefficient to silk fibroin: ['friction coefficient', 'pH', 'silk fibroin']
    
    Found path from hydrophobicity to silk fibroin: ['hydrophobicity', 'pH', 'silk fibroin']
    
    Found path from adhesion to silk fibroin: ['adhesion', 'ionic strength', 'silk fibroin']
    
    Found path from tetragonal structure to silk fibroin: ['tetragonal structure', 'X-ray diffraction', 'Silk fibroin-based bentonite composite', 'silk fibroin']
    
    Found path from tribological behavior to silk fibroin: ['tribological behavior', 'Scanning electron microscopy', 'Polyurethanes (PUs)', 'silk fibroin']
    
    Found path from surface energy to silk fibroin: ['surface energy', 'surface chemistry', 'nanocomposite membranes', 'silk fibroin']
    
    Found path from covalent bonds to silk fibroin: ['covalent bonds', 'adhesion', 'ionic strength', 'silk fibroin']
    
    Found path from Viscosity Measurements to silk fibroin: ['Viscosity Measurements', 'fluorine-containing polyimide', 'nitrogen', 'silk fibroin']
    
    Found path from efficient packing to silk fibroin: ['efficient packing', 'Imidazolium', 'Ionic liquid', 'hydrogen bonding', 'silk fibroin']
    
    Found path from cryopreservation techniques to silk fibroin: ['cryopreservation techniques', 'HSPC', 'reagents', 'alpha-amylase', 'pH', 'silk fibroin']
    
    Found path from biocompatibility to silk fibroin: ['biocompatibility', 'Eutectogels', 'silk fibroin']
    
    Found 13 paths out of 91 node pairs that include 'silk fibroin'
    
    \textbf{\hlgray{Please consider these knowledge graph relationships:}}
    
    Eutectogels exhibits biocompatibility. Eutectogels reinforced\_with silk fibroin. tribological behavior includes friction coefficient. tribological behavior influenced by surface chemistry. nanocomposite membranes has\_property surface chemistry. pH affects activity of alpha-amylase. pH affects hydrogen bonding. pH affects ionic strength. pH affects conformation of silk fibroin. pH affects adhesion. pH affects friction coefficient. silk fibroin enables hydrogen bonding. silk fibroin component of nanocomposite membranes. silk fibroin component of Silk fibroin-based bentonite composite. silk fibroin contains nitrogen. surface energy affects biocompatibility. surface energy affects hydrophobicity. surface energy increases adhesion. surface energy affects friction coefficient. friction coefficient changes to friction coefficient. friction coefficient has a positive relationship with adhesion. friction coefficient independent of surface energy. friction coefficient depends\_on pH. friction coefficient is a part of tribological behavior. surface chemistry includes surface energy. surface chemistry influences biocompatibility. surface chemistry affects hydrophobicity. surface chemistry includes friction coefficient. covalent bonds forms strong adhesion. Imidazolium improves efficient packing. Imidazolium type of cation Ionic liquid. hydrogen bonding increases adhesion. hydrogen bonding enhances dimensional stability of Polyimides. alpha-amylase affected by pH. HSPC affected by cryopreservation techniques. HSPC affected by reagents. X-ray diffraction confirmed by tetragonal structure. X-ray diffraction characterizes Silk fibroin-based bentonite composite. nitrogen affects friction coefficient. ionic strength affects adhesion. ionic strength affects adsorption of silk fibroin. fluorine-containing polyimide is a type of Polyimides. fluorine-containing polyimide stable in nitrogen. fluorine-containing polyimide characterized by X-ray diffraction. fluorine-containing polyimide characterized by Viscosity Measurements. fluorine-containing polyimide affects biocompatibility. Scanning electron microscopy shows hydrolytic degradation of Polyurethanes (PUs). Scanning electron microscopy characterizes morphology of Silk fibroin-based bentonite composite. Scanning electron microscopy used to analyze tribological behavior. reagents affects activity of alpha-amylase. hydrophobicity influences adhesion. hydrophobicity increases with increased hydrophobicity friction coefficient. hydrophobicity affected by pH. Polyimides tested in nitrogen. Polyimides possess adhesion. Polyimides has property high-temperature resistance. Ionic liquid is a replacement for reagents. Ionic liquid possess hydrogen bonding. Ionic liquid consists of Imidazolium. Ionic liquid reduces friction coefficient. Polyurethanes (PUs) is cross-linked with silk fibroin. adhesion is improved without covalent bonds. adhesion determined by ionic strength. adhesion affects friction coefficient.
    
  \end{mdframed}
  
  \caption{Creative GraphWeave agent with BFS algorithm with a semantic stop set to \textit{silk}. The agent extracts material property keywords from the Evaluator and maps them to the knowledge graph, searching for relational chains that explicitly involve silk fibroin. While many keyword pairs yield no valid connections, selected paths highlight the role of silk fibroin in mediating property relationships. The paths themselves uncover adjacent concepts and mechanistic linkages, which can be leveraged by the Engineer agent for hypothesis generation around silk fibroin-based materials.}
  \label{fig:silk_fibroin_paths}
\end{figure}

Figure~\ref{fig:engineer_BFS_semantic} presents an example output generated using the BFS algorithm with the semantic stop node \textit{Silk}. By forcing all keyword-pair paths to traverse through silk fibroin, the algorithm steers the hypothesis toward a silk-centric material design, proposing a silk fibroin matrix reinforced with titanium dioxide (TiO$_2$) nanoparticles and integrated into a eutectogel system. The hypothesis emphasizes silk fibroin's biocompatibility and hydrogen bonding capabilities as the foundational matrix properties, while TiO$_2$ nanoparticles contribute friction reduction and additional biocompatibility through knowledge graph paths connecting tribological behavior to surface chemistry. The incorporation of eutectogels addresses thermal stability requirements by leveraging graph connections between silk fibroin reinforcement and temperature-stable gel systems. The justification centers on strategies for exploiting silk's molecular interactions, particularly its role in stabilizing emulsions through hydrogen bonding and pH-dependent conformational changes, while balancing mechanical strength through TiO$_2$ reinforcement and thermal performance through the eutectogel component. This semantic stop approach demonstrates how user-defined constraints can guide hypothesis generation toward specific material families while still exploring complementary components through adjacent graph neighborhoods.

\subsection{Ablation Study}

To evaluate the contribution of each agent, we conducted a system-wide ablation study comparing progressively simplified configurations against the full multi-agent pipeline. The goal of this study is to demonstrate that our approach outperforms directly prompting the foundational LLM and that each added agent measurably improves the quality of the system’s output. Five configurations were evaluated, ranging from the simplest system, that is prompting the Engineer agent only, to the complete multi-agent system as described in this study. All evaluated systems which includes the Creative GraphWeave agent made use of the Shortest Simplest Path algorithm for traversing the knowledge graph. Table \ref{tab:ablation-configs} summarizes the five agentic system configurations evaluated in this study and the agents included in each.

\begin{table}[hb]
\centering
\caption{Summary of the five multi-agent system configurations evaluated in the ablation study.}
\label{tab:ablation-configs}
\begin{tabular}{lccccc}
\hline
\textbf{Configuration} & \textbf{Planner} & \textbf{Hybrid GraphWeave} & \textbf{Evaluator} & \textbf{Creative GraphWeave} & \textbf{Engineer} \\
\hline
All Components Enabled & \checkmark & \checkmark & \checkmark & \checkmark & \checkmark \\
Expanded Retrieval        & \checkmark & \checkmark & $\times$ & $\times$ & \checkmark \\
Direct Graph Exploration  & $\times$ & \checkmark & $\checkmark$ & \checkmark & \checkmark \\
Direct Retrieval          & $\times$ & \checkmark & $\times$ & $\times$ & \checkmark \\
LLM Only             & $\times$ & $\times$ & $\times$ & $\times$ & \checkmark \\

\hline
\end{tabular}
\end{table}

Each system was tested with the same prompt, and its outputs were evaluated in blind trials by GPT-5 configured for high reasoning effort and high verbosity \cite{openai2024gpt5card}.  Using powerful LLMs such as OpenAI’s GPT series for evaluation of outputs from agentic systems has been shown effective in prior studies \cite{ghafarollahi2025sciagents,ghafarollahi2025sparks,NEURIPS2023_91f18a12}. To capture different dimensions of performance, we define six evaluation criteria for the model to evaluate against: task decomposition, context enrichment, cross-subtask integration, deep reasoning, novelty, and source attribution. We select these criteria as they reflect essential capabilities for hypothesis generation in materials science and correspond to the distinct strengths of individual agents, thereby making it possible to evaluate how each agent contributes to or limits the overall framework. A summary of the six evaluation criteria, together with their descriptions provided to the evaluator LLM, is listed in Table \ref{tab:evaluation-criteria}.

\begin{table}[hb]
\centering
\caption{Summary of the six evaluation criteria used in the ablation study, with names and descriptions as provided to the evaluator LLM.}
\label{tab:evaluation-criteria}
\begin{tabular}{lp{11.25cm}}
\hline
\textbf{Criterion} & \textbf{Description} \\
\hline
Task Decomposition & Degree to which the system breaks the initial task into meaningful subtasks for more targeted reasoning and retrieval. \\
Context Enrichment & Extent and relevance of external information retrieved to enhance the model's understanding of the task. \\
Cross-Subtask Integration & Ability to integrate and reason across enriched contexts from different subtasks to develop more holistic insights. \\
Deep Reasoning & Explores relationships between concepts and generates novel insights. \\
Novelty & Originality and innovativeness of the proposed insights, hypotheses, or solutions. \\
Source Attribution & Clarity and consistency in referencing retrieved sources of information. \\
\hline
\end{tabular}
\end{table}

By systematically comparing these configurations under consistent evaluation criteria, the ablation study seeks to isolate the contribution of each agent, thereby demonstrating how the system improves incrementally as additional reasoning and retrieval components are introduced.

\begin{figure}[!t]
    \centering
    \includegraphics[width=0.75\linewidth]{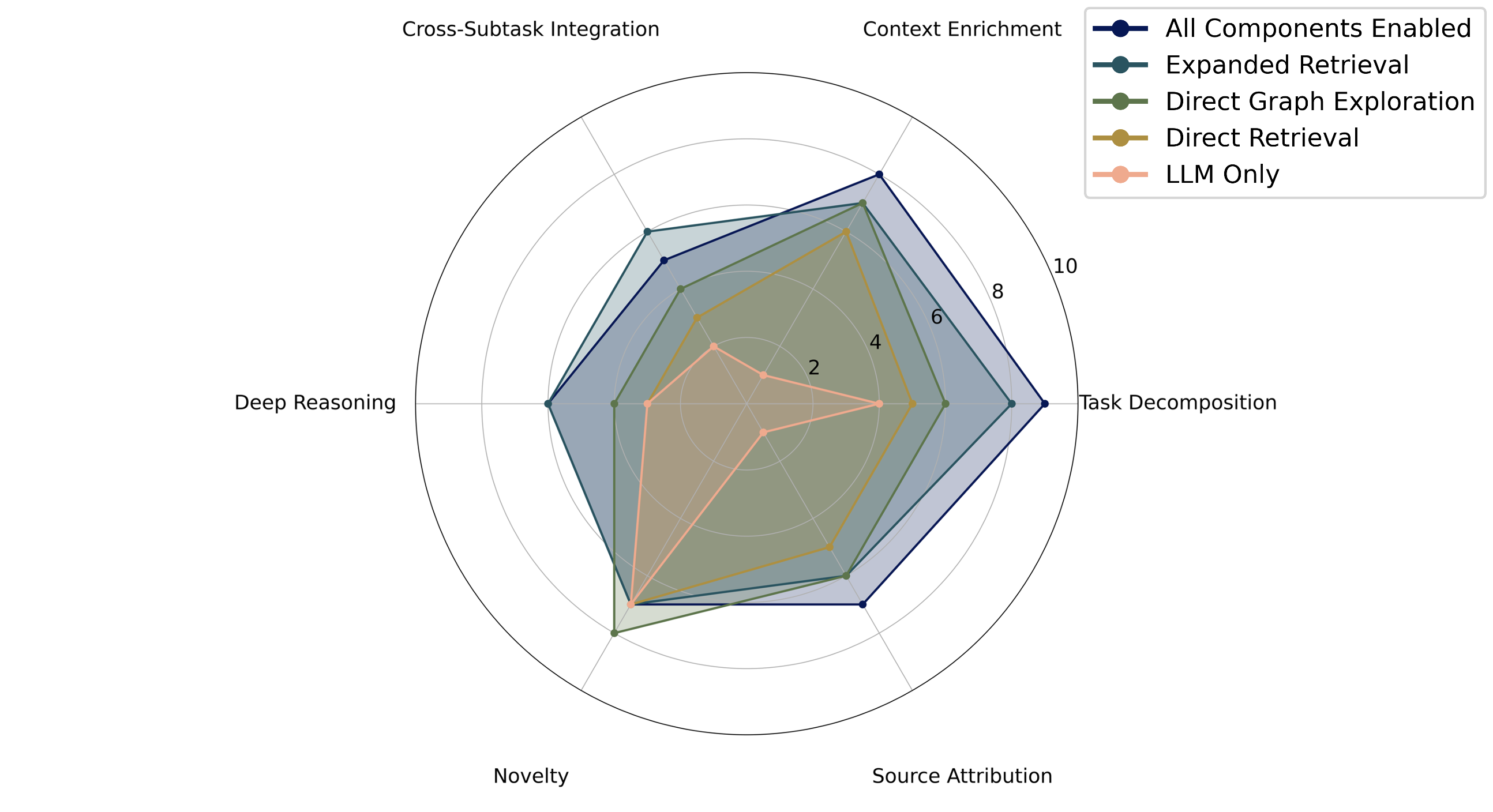}
    \caption{Ablation study outcomes shown as a spider chart, comparing the five system configurations across the six evaluation criteria. The full multi-agent pipeline achieves the highest overall performance. The corresponding scores, reported in Table \ref{tab:ablation-scores}, highlight the pipeline’s advantage and its ability to balance nuanced retrieval, relational reasoning, and traceable evidence.}
    \label{fig:spider}
\end{figure}

The results of the ablation study are presented in Table~\ref{tab:ablation-scores}, while Figure~\ref{fig:spider} presents the results of the ablation study in the form of a spider chart, comparing the results of the five configurations of the system across the six evaluation criteria. The full multi-agent pipeline achieves the highest overall performance, surpassing or matching all ablated configurations across all evaluation criteria, with the exception of cross-subtask integration and novelty. The results demonstrate that the full pipeline consistently outperforms its ablated variants, highlighting its ability to balance nuanced retrieval, relational reasoning, and traceable evidence.

\begin{table}[ht]
\centering
\caption{Scores for each system configuration across the six evaluation criteria, corresponding to the spider chart in Figure \ref{fig:spider}. The full multi-agent pipeline attains the highest overall performance. The results highlight how the complete system balances nuanced retrieval, relational reasoning, and traceable evidence more effectively than simplified configurations.}
\label{tab:ablation-scores}
\begin{tabular}{lcccccc}
\hline
\textbf{Configuration} & \makecell{\textbf{Task} \\ \textbf{Decomp.}} & \makecell{\textbf{Context} \\ \textbf{Enrich.}} & \makecell{\textbf{Cross-Subtask} \\ \textbf{Int.}} & \makecell{\textbf{Deep} \\ \textbf{Reasoning}} & \makecell{\textbf{Novelty}} & \makecell{\textbf{Source} \\ \textbf{Attr.}} \\
\hline
All Components Enabled    & 9 & 8 & 5 & 6 & 7 & 7 \\
Expanded Retrieval        & 8 & 7 & 6 & 6 & 7 & 6 \\
Direct Graph Exploration  & 6 & 7 & 4 & 4 & 8 & 6 \\
Direct Retrieval          & 5 & 6 & 3 & 3 & 7 & 5 \\
LLM Only                  & 4 & 1 & 2 & 3 & 7 & 1 \\
\hline
\end{tabular}
\end{table}

The Direct Graph Exploration setup exceeds the All-Components-Enabled configuration in novelty. The slightly lower novelty score of the full pipeline reflects the more grounded nature of its generated hypotheses, which are based on a broader and more comprehensive retrieved knowledge base than those from Direct Graph Exploration. Both systems use the Creative GraphWeave Agent, which allows for more exploratory reasoning than the other ablations that do not include this component. While the LLM cOnly system shows strong creativity, its outputs are often speculative and not guided by external knowledge. In contrast, the Direct and Expanded Retrieval configurations achieve a better balance between creativity and reliability, generating hypotheses that are imaginative but still supported by retrieved information.

Each agent in the system is designed to enhance a specific capability. For example, the Hybrid GraphWeave agent functions as an evidence finder, retrieving and synthesizing relevant information for each subtask to improve context enrichment, while the Planner acts as a design-oriented thinker that decomposes the user’s query into subquestions, thereby strengthening task decomposition. Consequently, the Direct Retrieval and Direct Graph Exploration systems both outperform the LLM Only baseline in context enrichment. The Expanded Retrieval system benefits further from the Planner’s influence on task decomposition and thus from richer context enrichment, since multiple subtasks broaden the pool of retrieved evidence. As a result, it surpasses the two previously discussed systems in both task decomposition and context enrichment.

Meanwhile, the Evaluator and Creative GraphWeave agents work together to strengthen cross-task reasoning. The Evaluator distills key design context and keywords from the retrieved information, and Creative GraphWeave then performs heuristic graph traversal to connect these keywords within the Material Properties Knowledge Graph. This process yields a subgraph that uncovers relationships and cross-domain linkages. As a result, the Direct Graph Exploration system outperforms Direct Retrieval in deep reasoning, indicating that Creative GraphWeave enables exploration of interconnected concepts across retrieved knowledge. However, we also find that simply adding Creative GraphWeave is insufficient for cross-subtask integration. The Planner is essential for decomposing the initial query into subtasks, which then allows Creative GraphWeave to link concepts across those subtasks and integrate their findings effectively. This is supported by Expanded Retrieval outperforming all other systems in cross-subtask integration, and matching the All Components Enabled system in deep reasoning.

Finally, in all configurations the Engineer agent integrates the evidence gathered across the multi-agent pipeline to generate hypotheses grounded in the available knowledge representation. As components are gradually removed from the pipeline, the Engineer’s performance systematically declines, with the most simplified setups performing the weakest. Furthermore, systems that incorporate external sources during retrieval achieve higher source-attribution scores, which is important when developing hypothesis based on traceable ideas and information. Expanded Retrieval outperforms Direct Retrieval by gathering a larger set of sources during its context-enrichment stage, whereas Direct Graph Exploration matches Expanded Retrieval by collecting additional sources during graph exploration. The All Components Enabled system achieves the best performance, leveraging both the Planner and Hybrid GraphWeave to retrieve a broad set of sources and the Creative GraphWeave agent to connect them across subtasks. As expected, the LLM Only system scores poorly in source attribution because it does not collect any sources, its answer relies solely on the initial query and self-generated content. Together, these results highlight the complementary contributions of each agent to the system’s depth, creativity, and credibility.

\begin{figure}[htbp]
    \centering
    \includegraphics[width=0.7\linewidth]{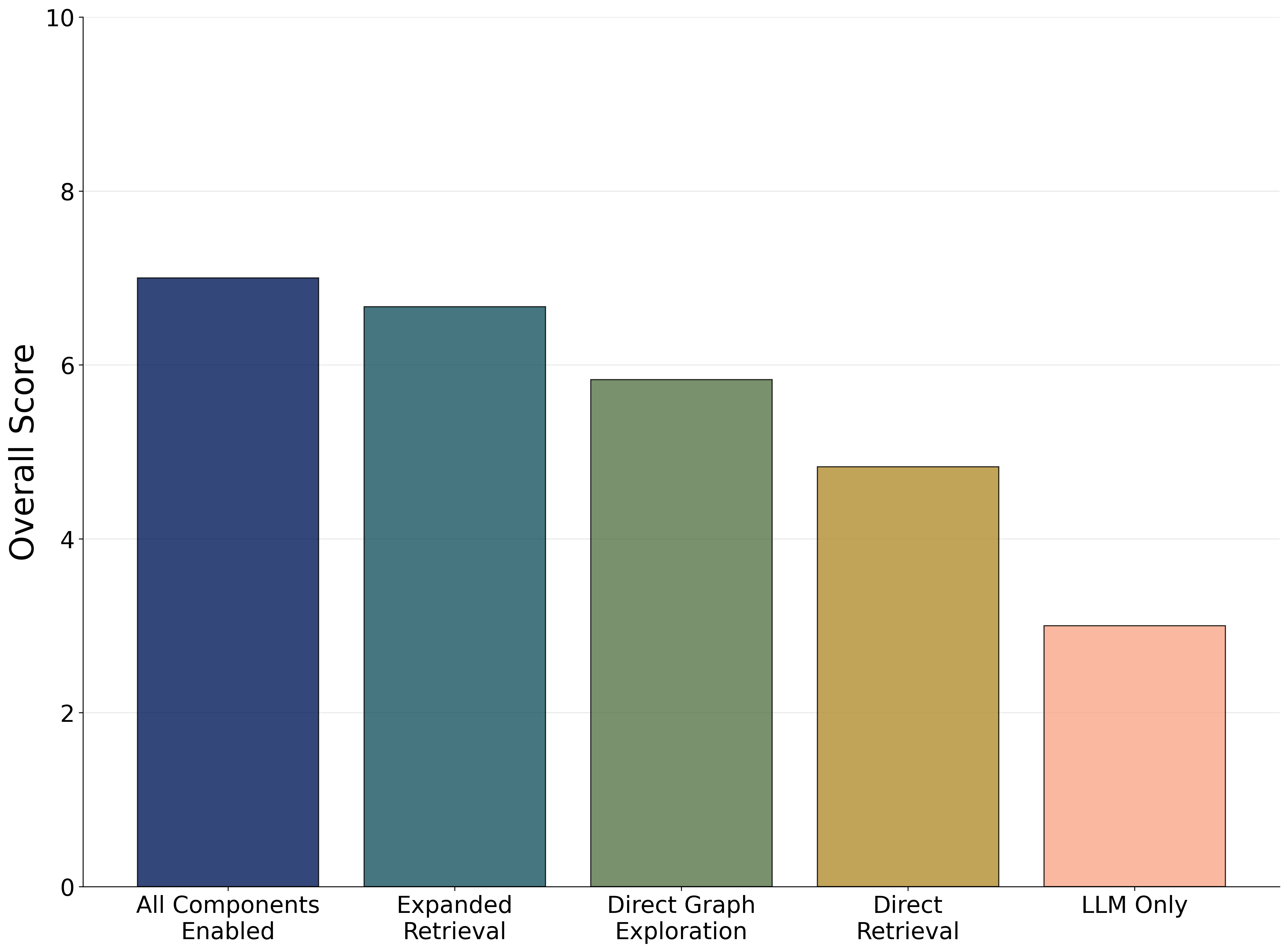}
    \caption{Overall scores for each system configuration, averaged across the six evaluation criteria. The results show a consistent decline in performance as components are removed, with the largest gap between the full multi-agent pipeline and the minimal LLM Only setup. Intermediate configurations highlight the contributions of individual agents, confirming that each agent adds value to the framework, and that the complete five-agent system achieves the most balanced and robust performance for hypothesis generation.}
    \label{fig:bar}
\end{figure}

In summary, Figure \ref{fig:bar} presents the overall scores for each configuration, averaged across the six evaluation criteria, and illustrates the consistent decline in performance as components are removed. The largest gap is observed between the full pipeline and the minimal LLM Only setup, underscoring the importance of layering complementary reasoning and retrieval modules. The intermediate configurations further demonstrate the specific value of individual agents where removing the Planner reduces nuance, removing Hybrid GraphWeave limits contextual grounding, and removing Creative GraphWeave diminishes relational exploration. Collectively, these results establish that each agent contributes to the system’s overall performance, and that the complete five-agent framework provides the most robust and well-rounded approach to hypothesis generation.

\begin{table}[h!]
\centering
\begin{tabular}{|l|r|}
\hline
\textbf{Keyword} & \textbf{Result Count} \\
\hline
Impermeability of small organic molecules & 4 \\
Impermeability of small molecules & 77 \\
Low absorption of flavors & 563 \\
Low absorption of small molecules & 5,569 \\
Low diffusion of small molecules & 3,558 \\
Low surface energy leads to low adsorption & 7,069 \\
Chemically inert & 4,258 \\
Compatible with Sterilization & 191 \\
Sterilization Safe & 1,636 \\
Biocompatible and Non-reactive & 14 \\
Low adsorption of proteins & 15,075 \\
Biocompatible and Non-adherence & 5 \\
High light transmission and low refractive index & 653 \\
Transparent polymers & 23,338 \\
Cold and cryogenic & 5,071 \\
Broad Temperature Stability & 10,747 \\
Flexible at low temperatures & 16,795 \\
Very high chemical resistance & 6,819 \\
Very high thermal resistance & 7,319 \\
Extreme chemical resistance & 2,110 \\
Extreme thermal resistance & 3,138 \\
\textit{High purity} & 112,314 \\
\textit{High purity material} & 41,445 \\
Low extractable and leachable & 83 \\
Chemical resistance to strong acid & 2,528 \\
Chemical resistance to organic solvents & 1,935 \\
Chemical resistance to methanol & 1,702 \\
Chemical resistance to acetone & 813 \\
Chemical resistance to DMSO & 229 \\
Chemical resistance to protein denaturants & 63 \\
Chemical resistance to guanidine hydrochloride & 50 \\
Chemical resistance to guanidine thiocyanate & 1 \\
Chemical resistance to immunoassay reagents & 6 \\
Chemical resistance to luminol & 15 \\
Chemical resistance to acridinium esters & 0 \\
Resistance to acridinium esters & 6 \\
Resistance to AMPPD & 0 \\
Chemical resistance to enzymes & 8,842 \\
Chemical resistance to alkaline phosphatase & 245 \\
Chemical resistance to horseradish peroxide & 14 \\
Chemical resistance to fluorescent dyes & 196 \\
\textit{Low surface energy} & 346,417 \\
\textit{Low surface energy material} & 152,129 \\
Low coefficient of friction material & 14,309 \\
Survive lamination process & 15 \\
Lamination compatible & 183 \\
Lamination flexible & 705 \\
Lower environmental impact material & 35,289 \\
Laminates well to silicone & 41 \\
Laminates to silicone & 266 \\
Laminates no primer & 23 \\
Flexible, no kinking as it bends & 2 \\
No kink bends & 398 \\
Flexible no kink bends & 24 \\
\hline
\end{tabular}
\caption{Polymer experts identified pertinent PFAS material property keywords, together with the corresponding Web of Science Core Collection manuscript counts retrieved through the keyword search.}
\captionsetup{skip=5pt}
\label{tab:pfas_keywords}
\end{table}

\section{Conclusion and Future Directions}
\label{sec:conclusion}
The multi-agent pipeline developed in this work emulates and extends beyond human processes of hypothesis generation by synthesizing information around a design problem and extrapolating to adjacent domains of knowledge. We make use of targeted injections within in-context learning of LLMs which can critically shape model behavior and guide generated output. In this method, an injection refers to the deliberate inclusion of structured knowledge, constraints, or illustrative examples into the model’s immediate context, thereby ensuring that its reasoning process remains aligned with user objectives. By curating the information presented during inference, researchers can direct the model toward recognizing patterns or relationships outside the LLMs training data that might otherwise remain hidden. This approach is particularly valuable in complex domains such as materials discovery, where the design space is vast and unconstrained. Well-designed injections allow the model to focus on salient properties, standards, and experimental protocols, thereby improving both the reliability and interpretability of its outputs. Through this mechanism, in-context injections transform the model from a generic language processor into a specialized assistant capable of supporting scientific problem solving.

We demonstrate this approach in the search for sustainable alternatives to PFAS materials, where in-context content is dynamically tuned using external knowledge bases and tools accessible to agents on demand. In particular, we integrate knowledge graphs of PFAS and their intrinsic material properties, enabling agents to navigate a broader knowledge space and apply graph traversal algorithms to uncover latent connections embedded in the graph but not yet explicitly recognized in the literature. We stipulate that the materials and design hypotheses discussed in the PFAS use case are presented to illustrate the behavior and capabilities of the multi-agent system and to highlight early-stage development potential, rather than to prescribe implementation-ready material substitutions. It is important to note that the study and examples are used to illustrate the models rather than being final solutions, where further work is necessary, including experimental validation.

By design, our knowledge graphs currently generate hypotheses rooted in an academic corpus, leading to proposed composites that emphasize scientifically novel or exotic materials (e.g., graphene oxide, cellulose nanocrystals, lignin blends) but are not always industrially viable. To overcome this limitation, future work will expand the corpus to include diverse sources such as patents and technical datasheets, which capture industry-relevant testing protocols, performance standards, and material metrics. Incorporating these sources will enhance design-problem formulation and facilitate the identification of key design parameters, enabling tighter alignment with corresponding nodes in the global knowledge graph. This expansion will, in turn, improve the accuracy and realizability of the predicted property profiles, ensuring closer conformity with the design objectives and performance characteristics of PFAS. Further, wider integration of external datasets will broaden the range of traversable paths within the graph, enriching the substrate from which the Engineer agent can derive candidate designs. Providing the agentic system with greater access to validation tools would establish stronger feedback loops, improving the generated hypotheses and guiding iterative refinement of material designs. 

Currently, GraphAgents relies on pre-computed, static knowledge graphs. A logical next step is to enable agents to construct their own reasoning substrates dynamically during inference. Future iterations could integrate \textit{in-situ} knowledge expansion mechanisms, such as the Graph-PReFLexOR framework~\cite{Buehler2025GraphPRefLexOR,Buehler2025Chaos}, allowing the system to build temporary `scaffolding' graphs to bridge distant domains only when necessary. Furthermore, coupling this creative engine with autonomous synthesis labs would close the loop, transforming these digital hypotheses into physical reality without human intervention.

\section{Materials and Methods}
\label{sec:methods}
\subsection{Knowledge Graphs for Enhanced Agent Reasoning}

Two large-scale knowledge graphs were developed for this study: one PFAS-Specific Knowledge Graph derived from PFAS-related scientific papers, and one broad Material Properties Knowledge Graph derived from full abstracts collected using a curated set of PFAS-related material property keywords. The graphs’ sizes in terms of nodes and edges are listed in Table \ref{tab:knowledge graph-sizes}. In both cases, nodes correspond to entities such as materials, properties, or contextual features, while edges capture the relationships between them. The PFAS-Specific Knowledge Graph is designed for depth, focusing on the specific rationale and performance requirements of PFAS in real-world applications. The broader Material Properties Knowledge Graph, by contrast, provides breadth, organizing knowledge around 51 property categories that characterize PFAS. Together, these graphs provide a dual substrate: one enabling targeted exploration within PFAS-specific literature, and the other allowing agents to traverse adjacent domains for more creative and cross-disciplinary hypothesis generation.

\begin{table}[H]
\centering
\caption{Sizes of the two knowledge graphs used in this study.}
\label{tab:knowledge graph-sizes}
\begin{tabular}{lcc}
\hline
\textbf{Knowledge Graph} & \textbf{Nodes} & \textbf{Edges} \\
\hline
PFAS-Specific            & 144,380        & 458,643 \\
Material Properties      & 315,020        & 773,541 \\
\hline
\end{tabular}
\end{table} 

The knowledge graphs are constructed through a systematic pipeline that transforms unstructured scientific literature into structured graph representations. First, a collected corpus of scientific literature is converted into machine readable markdown files and segmented into chunks, each kept within the context length limits of a deployed LLM. Then, each chunk is processed by this LLM to extract semantic triplets, that is two nodes corresponding to entities and an edge representing the entities' relationship, which forms a local subgraph of triplets. Repeating this process across all chunks produces multiple local subgraphs that are subsequently merged into a single large-scale graph. During merging, nodes identified as sufficiently similar based on textual embeddings are consolidated to avoid duplication. This pipeline facilitates generation of knowledge graphs that capture both the breadth of the source corpora and the interconnections across domains. 

\subsection{Selected Models, Hosting, and Supporting Libraries}

The open source LLM \texttt{meta-llama/Llama-3.3-70B-Instruct} \cite{llama3modelcard} was used for both knowledge graph generation and as the agents' language backbone in the multi-agent framework. The model was hosted locally with \texttt{llama.cpp} \cite{gerganov2023llamacpp}, enabling efficient inference and secure deployment through an OpenAI-compatible API schema. To optimize performance, the server was configured with full GPU acceleration, tensor splitting across multiple devices, and an extended context length of 40,000 tokens to accommodate long scientific prompts. The model was deployed in a quantized Q4 format, reducing memory requirements while preserving strong performance, and was loaded directly from internal infrastructure ensuring all data remained on local servers.

This setup provided a lightweight yet flexible inference interface, supporting scalable integration with both the knowledge graph generation pipelines and the multi-agent framework. For knowledge graph generation, the extended context length and efficient local hosting enabled the model to process large volumes of scientific text and extract entities and relations effectively. In the multi-agent framework, the same setup enabled fast and reliable inference with enough contextual capacity to support task decomposition, grounding, and hypothesis generation based on the retrieved information supplied in the context-enriched prompts.

Several open-source Python packages were employed to implement the knowledge graph construction and retrieval pipeline in this study. For text embeddings, the \texttt{sentence-transformers} library \cite{reimers-2019-sentence-bert} was used to load and apply the open source \texttt{nomic-ai/nomic-embed-text-v1.5} model \cite{nussbaum2024nomic}, which produces semantic vector representations of text for similarity search and retrieval-augmented generation. This model was selected for its strong performance on both short- and long-context embedding benchmarks, as well as its open-source availability, which makes it well suited for large-scale knowledge graph embedding and retrieval.

The \texttt{NetworkX} package \cite{SciPyProceedings_11} was used to manipulate knowledge graph data structures, including managing node and edge attributes. For embedding storage and retrieval, the \texttt{Chroma} library \cite{chroma} was adopted for the vector database. Its persistent client enables local storage of embedding collections and supports efficient similarity queries using cosine distance. The open-source \texttt{AutoGen} framework \cite{wu2023autogenenablingnextgenllm} was used to facilitate the multi-agent pipeline, enabling both automated LLM-driven generation and tool usage, such as querying the knowledge graphs.

\subsection{In-House Tools for Knowledge Graph Retrieval}

Several custom tools were used and developed to enable agents to interact with the generated knowledge graphs beyond simple text retrieval. Our \texttt{GraphWeave} method extracts subgraphs linked to document identifiers, collects connected nodes, and summarizes relationships into structured text for agent interpretation. A core component is the \texttt{keyword-to-node} mapping tool, which identifies the closest matching graph nodes to input keywords via embedding comparisons. 

As discussed above, we utilized the \texttt{GraphWeave} method in two tooled approaches to enable the Hybrid GraphWeave and Creative GraphWeave agents. The Hybrid GraphWeave approach combines evidence from both the raw text corpus stored in the \texttt{Chroma} library \cite{chroma} and the knowledge graph by embedding a user query, retrieving top-matching text chunks by cosine-similarity, and linking these chunks to corresponding subgraphs. This dual process ensures that retrieval is grounded in direct textual evidence while also leveraging relational context from the graph. In contrast, the Creative GraphWeave approach focuses on exploratory reasoning within the knowledge graph itself, mapping query keywords to nodes through the \texttt{keyword-to-node} tool, relaxing directional constraints, and applying pathfinding algorithms to assemble subgraphs that expose previously unseen latent structural connections in the knowledge graph. 

Finally, a suite of heuristic graph traversal tools was implemented to extend reasoning across domains. These include Breadth-First Search (BFS) with Semantic Stop, Depth-First Search (DFS), Shortest Simple Path, and Top-$N$ Shortest Simple Path algorithms. These traversal tools allow for varying the enriched context that agents can leverage to propose more holistic and creative hypotheses.

\subsection{Corpus Development}
The corpus for the PFAS-Specific Knowledge Graph and the Material Properties Knowledge Graph was assembled using searches in the Web of Science Core Collection \cite{clarivate2025}. For the PFAS-Specific Knowledge Graph, the query term was ``Per- and polyfluoroalkyl substances''. This search yielded 4,824 results on February 18, 2025, of which 4,716 articles were successfully retrieved, corresponding to a yield of $\approx 97.8\%$. Articles that could not be obtained were primarily unavailable due to missing or inaccurate metadata. Full-text manuscripts were collected through a combination of publisher-provided application programming interfaces and manual scraping with permission. 

The Material Properties Knowledge Graph was constructed from the union of search results generated using each of the 51 property keywords listed verbatim that are non-italicized in Table~\ref{tab:pfas_keywords} Polymer scientists designated the italicized terms “high purity,” “high purity material,” “low surface energy,” and “low surface energy material” as key PFAS material properties; however, they were not incorporated into the knowledge graph due to the prohibitively high volume of manuscripts retrieved.

In contrast to the PFAS-specific collection, full-text acquisition for the Material Properties Knowledge Graph was limited by the sheer volume of papers, the broad distribution of publishers and their varying access policies, and API restrictions that rendered manual scraping impractical. To address this and maximize both coverage and diversity, we leveraged the Web of Science functionality to retrieve abstracts and metadata in batches of 1,000 records. These abstracts constitute the corpus for the Material Properties Knowledge Graph. The final collection, completed on April 21, 2025, contained 160,495 abstracts; however, the Material Properties Knowledge Graph used for multi-agent experimentation was from a random sample of 63,222 abstracts, representing a practical limit given the computational expense of graph merging, traversal, and information retrieval at this scale. We found this size to be sufficient for our Evaluator keywords to identify matches within the knowledge graph. As a preliminary validation of this multiagent pipeline, we observe that even at this final scale, the model demonstrates creativity and yields distinctive outcomes across graph traversal algorithms. The total number of abstracts originally retrieved for each material property keyword is reported in Table~\ref{tab:pfas_keywords}. Future work could improve computational and retrieval efficiency to enable larger-scale expansion and utility of the knowledge graph.

\subsection{Knowledge Graph Construction Pipeline}

We construct the knowledge graph through an established pipeline \cite{buehler_accelerating_2024} that first extracts graph fragments within each paper and then merges them across papers.

\subsubsection{Intra-Paper Extraction}
The retrieved PDFs are first converted to Markdown (\texttt{.md}) using \texttt{marker-pdf} \cite{marker-github}, which preserves core text structure, including section headers, the positions of tables and figures, and in-text reference labels, therefore facilitating downstream LLM-based knowledge extraction. We then segment the text into sections  to improve contextual resolution while reducing token usage. For each section, we perform multi-pass LLM distillation with structured outputs to produce a \emph{graph fragment}, essentially a list of triples, represented as a set of nodes and directed edges between nodes. The extraction uses the following schema in python:

\begin{verbatim}
    class Node(BaseModel):
        id: str
        type: str
            
    class Edge(BaseModel):
        source: str
        target: str
        relation: str
            
    class KnowledgeGraph(BaseModel):
        nodes: List[Node]
        edges: List[Edge]
\end{verbatim}

Then, since terminology within the same paper is highly correlated, we can directly compose all fragments from the same paper into a single knowledge graph, where the identical nodes will connect fragments within a paper.

\subsubsection{Cross-Paper Alignment and Merge}
To handle semantic variations in word use across our cross-domain corpus, we compute node embeddings using a \emph{nomic} sentence embedding model for all nodes so that we can compare them in a high-dimensional vector space; nodes with small angular distance, that is, high cosine similarity (Eq.~\ref{eq:1}) greater than 0.95, are treated as semantically similar and merged. For two nodes with embeddings $\mathbf{v}_i$ and $\mathbf{v}_j$, their cosine similarity is
\begin{equation}
\operatorname{cos\_sim}(\mathbf{v}_i,\mathbf{v}_j)
= \frac{\mathbf{v}_i^\top \mathbf{v}_j}{\lVert \mathbf{v}_i\rVert \,\lVert \mathbf{v}_j\rVert}.
\label{eq:1}
\end{equation}

Following Eq.~\ref{eq:1}, we merge similar node pairs by keeping the higher-degree node and removing the other. This step is essential at our corpus scale, preserving core information while reducing the memory footprint. Using this merging strategy, we iteratively integrate each paper’s component graph into the core knowledge graph, merging duplicates via cosine similarity, and finally produce the large knowledge graphs used in this study. 

\clearpage 
\section*{Acknowledgments}
\label{sec:acknowledgements} 
This research was supported by Saint-Gobain (Northboro Research and Development Center, United States). 
I.A.S. acknowledges that this material is based upon work supported by the U.S. Department of Energy, Office of Science, Office of Advanced Scientific Computing Research, under Award Number DE-SC0026073. T.P.H. acknowledges that this material is based upon work supported by Aker Scholarship.
We acknowledge valuable discussions with Lalitha Raghavan and Amir Kazemi-Moridani (both Saint-Gobain), whose contributions in identifying key PFAS material property keywords that shaped our corpus, maintaining the industrial relevance of the project, and guiding prospective experimental directions were instrumental to this work.

\section*{Author Contributions}
\label{sec:contributions}

M.J.B. supervised the project and led the research design and development. I.A.S. collected the data corpus, conceived the multi-agent framework and inference code, developed application of graph traversal methods, and carried out data analysis. T.P.H. conducted ablation studies on the multi-agent framework, performed the associated analyses, and contributed to data analysis. Y.-C.H. generated the knowledge graphs and performed data analysis. I.A.S. prepared the initial draft of the manuscript with contributions from T.P.H. All authors contributed to editing and finalizing the manuscript.

\section*{Competing Interests}
\label{sec:competingInterests}
The authors declare no competing interests.

\section*{Code and Data Availability}
\label{sec:codeAvailability}
All code, protocols, datassets, and notebooks developed in this study are available at: \url{https://github.com/lamm-mit/GraphAgents} . This repository also hosts supplementary files with the full agent outputs referenced in this manuscript.
\bibliographystyle{naturemag}

\bibliography{references,zotero_references}

@misc{openai2024gpt5card,
      title={{GPT-5 System Card}}, 
      author={OpenAI},
      year={2025},
      month={August},
      url={https://openai.com/index/gpt-5-system-card/},
}

@article{ghafarollahi2025sciagents,
  author       = {Alireza Ghafarollahi and Markus J. Buehler},
  title        = {SciAgents: Automating Scientific Discovery Through Bioinspired Multi-Agent Intelligent Graph Reasoning},
  journal      = {Advanced Materials},
  volume       = 37,
  number       = 22,
  year         = 2025,
  month        = 6,
  doi          = {10.1002/adma.202413523}
}

@article{ghafarollahi2025sparks,
  author       = {Alireza Ghafarollahi and Markus J. Buehler},
  title        = {Sparks: Multi-Agent Artificial Intelligence Model Discovers Protein Design Principles},
  journal      = {arXiv preprint arXiv:2504.19017},
  year         = 2025,
  month        = 4,
  doi          = {10.48550/arXiv.2504.19017}
}

@inproceedings{Vaswani2017attention,
 author = {Vaswani, Ashish and Shazeer, Noam and Parmar, Niki and Uszkoreit, Jakob and Jones, Llion and Gomez, Aidan N and Kaiser, \L ukasz and Polosukhin, Illia},
 booktitle = {Advances in Neural Information Processing Systems},
 editor = {I. Guyon and U. Von Luxburg and S. Bengio and H. Wallach and R. Fergus and S. Vishwanathan and R. Garnett},
 pages = {},
 publisher = {Curran Associates, Inc.},
 title = {Attention is All you Need},
 url = {https://proceedings.neurips.cc/paper_files/paper/2017/file/3f5ee243547dee91fbd053c1c4a845aa-Paper.pdf},
 volume = {30},
 year = {2017}
}

@misc{llama3modelcard,
  title={{Llama 3 Model Card}},
  author={AI@Meta},
  year={2024},
  url = {https://github.com/meta-llama/llama3/blob/main/MODEL_CARD.md}
}

@misc{gerganov2023llamacpp,
  author       = {Georgi Gerganov},
  title        = {{llama.cpp: Efficient LLM inference in C/C++}},
  year         = {2023},
  howpublished = {\url{https://github.com/ggml-org/llama.cpp}},
  note         = {MIT License}
}

@inproceedings{NEURIPS2023_91f18a12,
 author = {Zheng, Lianmin and Chiang, Wei-Lin and Sheng, Ying and Zhuang, Siyuan and Wu, Zhanghao and Zhuang, Yonghao and Lin, Zi and Li, Zhuohan and Li, Dacheng and Xing, Eric and Zhang, Hao and Gonzalez, Joseph E and Stoica, Ion},
 booktitle = {Advances in Neural Information Processing Systems},
 editor = {A. Oh and T. Naumann and A. Globerson and K. Saenko and M. Hardt and S. Levine},
 pages = {46595--46623},
 publisher = {Curran Associates, Inc.},
 title = {Judging LLM-as-a-Judge with MT-Bench and Chatbot Arena},
 url = {https://proceedings.neurips.cc/paper_files/paper/2023/file/91f18a1287b398d378ef22505bf41832-Paper-Datasets_and_Benchmarks.pdf},
 volume = {36},
 year = {2023}
}

@inproceedings{reimers-2019-sentence-bert,
  title = "Sentence-BERT: Sentence Embeddings using Siamese BERT-Networks",
  author = "Reimers, Nils and Gurevych, Iryna",
  booktitle = "Proceedings of the 2019 Conference on Empirical Methods in Natural Language Processing",
  month = "11",
  year = "2019",
  publisher = "Association for Computational Linguistics",
  url = "https://arxiv.org/abs/1908.10084",
}

@misc{nussbaum2024nomic,
      title={Nomic Embed: Training a Reproducible Long Context Text Embedder}, 
      author={Zach Nussbaum and John X. Morris and Brandon Duderstadt and Andriy Mulyar},
      year={2024},
      eprint={2402.01613},
      archivePrefix={arXiv},
      primaryClass={cs.CL}
}

@InProceedings{SciPyProceedings_11,
  author =       {Aric A. Hagberg and Daniel A. Schult and Pieter J. Swart},
  title =        {Exploring Network Structure, Dynamics, and Function using NetworkX},
  booktitle =   {Proceedings of the 7th Python in Science Conference},
  pages =     {11 - 15},
  address = {Pasadena, CA USA},
  year =      {2008},
  editor =    {Ga\"el Varoquaux and Travis Vaught and Jarrod Millman},
}

@misc{chroma,
  author       = {Chroma},
  title        = {{Chroma - the open-source embedding database}},
  howpublished = {GitHub repository},
  year         = {2025},
  note         = {Apache 2.0 License},
  url          = {https://github.com/chroma-core/chroma}
}

@article{Buehler2024XLoRA,
  title        = {X-LoRA: Mixture of low-rank adapter experts, a flexible framework for large language models with applications in protein mechanics and molecular design},
  author       = {Buehler, Eric L. and Buehler, Markus J.},
  journal      = {APL Machine Learning},
  year         = {2024},
  volume       = {2},
  pages        = {026119},
  doi          = {10.1063/5.0203126},
  url          = {https://doi.org/10.1063/5.0203126}
}

@article{Buehler2025GraphPRefLexOR,
  author    = {Buehler, Markus J.},
  title     = {In Situ Graph Reasoning and Knowledge Expansion Using {G}raph-{P}{R}ef{L}ex{O}{R}},
  journal   = {Advanced Intelligent Discovery},
  year      = {2025},
  pages     = {202500006},
  doi       = {10.1002/aidi.202500006},
  url       = {https://doi.org/10.1002/aidi.202500006}
}

@article{Buehler2025chaos,
  title={Self-organizing graph reasoning evolves into a critical state for continuous discovery through structural--semantic dynamics},
  author={Buehler, Markus J.},
  journal={Chaos},
  volume={35},
  number={11},
  pages={113117},
  year={2025},
  doi={10.1063/5.0272412},
  url={https://doi.org/10.1063/5.0272412},
  publisher={AIP Publishing}
}

@misc{wu2023autogenenablingnextgenllm,
      title={AutoGen: Enabling Next-Gen LLM Applications via Multi-Agent Conversation}, 
      author={Qingyun Wu and Gagan Bansal and Jieyu Zhang and Yiran Wu and Beibin Li and Erkang Zhu and Li Jiang and Xiaoyun Zhang and Shaokun Zhang and Jiale Liu and Ahmed Hassan Awadallah and Ryen W White and Doug Burger and Chi Wang},
      year={2023},
      eprint={2308.08155},
      archivePrefix={arXiv},
      primaryClass={cs.AI},
      url={https://arxiv.org/abs/2308.08155}, 
}

@misc{marker-github,
  author       = {Vik Paruchuri},
  title        = {Marker},
  year         = {2025},
  howpublished = {\url{https://github.com/datalab-to/marker}},
  note         = {GitHub repository; accessed 2025-09-22}
}

@article{Buehler2023MechGPTModalities,
    title = {{MechGPT, a language-based strategy for mechanics and materials modeling that connects knowledge across scales, disciplines and modalities}},
    year = {2023},
    journal = {Appl. Mech. Rev.},
    author = {Buehler, Markus J.},
    month = {10},
    url = {https://doi.org/10.1115/1.4063843},
    arxivId = {2310.10445}
}

@book{mac1998categories,
  title={Categories for the working mathematician},
  author={Mac Lane, Saunders},
  volume={5},
  year={1998},
  publisher={Springer Science \& Business Media}
}

@incollection{marquis_category_2023,
  author       = {Marquis, Jean-Pierre},
  title        = {Category Theory},
  booktitle    = {The {Stanford Encyclopedia of Philosophy} (Fall 2023 Edition)},
  editor       = {Zalta, Edward N. and Nodelman, Uri},
  publisher    = {Metaphysics Research Lab, Stanford University},
  year         = {2023},
  url          = {https://plato.stanford.edu/archives/fall2023/entries/category-theory/}
}

@misc{clarivate2025,
  author       = {{Clarivate™}},
  title        = {Certain data included herein are derived from Clarivate™ (Web of Science™). © Clarivate 2025. All rights reserved.},
  howpublished = {Web of Science™},
  year         = {2025}
}

@article{stewart_molecular_2025,
	title = {Molecular analysis and design using generative artificial intelligence \textit{via} multi-agent modeling},
	volume = {10},
	issn = {2058-9689},
	url = {https://xlink.rsc.org/?DOI=D4ME00174E},
	doi = {10.1039/D4ME00174E},
	language = {en},
	number = {4},
	urldate = {2026-02-06},
	journal = {Molecular Systems Design \& Engineering},
	author = {Stewart, Isabella and Buehler, Markus J.},
	year = {2025},
	pages = {314--337},
}

@misc{stewart_higher-order_2026,
	title = {Higher-{Order} {Knowledge} {Representations} for {Agentic} {Scientific} {Reasoning}},
	copyright = {Creative Commons Attribution Non Commercial No Derivatives 4.0 International},
	url = {https://arxiv.org/abs/2601.04878},
	doi = {10.48550/ARXIV.2601.04878},
	urldate = {2026-02-06},
	publisher = {arXiv},
	author = {Stewart, Isabella A. and Buehler, Markus J.},
	year = {2026},
	note = {Version Number: 1},
}

@article{giesa_category_2012,
	title = {Category {Theory} {Based} {Solution} for the {Building} {Block} {Replacement} {Problem} in {Materials} {Design}},
	volume = {14},
	copyright = {http://onlinelibrary.wiley.com/termsAndConditions\#vor},
	issn = {1438-1656, 1527-2648},
	url = {https://advanced.onlinelibrary.wiley.com/doi/10.1002/adem.201200109},
	doi = {10.1002/adem.201200109},
	language = {en},
	number = {9},
	urldate = {2025-10-06},
	journal = {Advanced Engineering Materials},
	author = {Giesa, Tristan and Spivak, David I. and Buehler, Markus J.},
	month = sep,
	year = {2012},
	pages = {810--817},
}

@misc{markovic_optimizing_2025,
	title = {Optimizing the {Interface} {Between} {Knowledge} {Graphs} and {LLMs} for {Complex} {Reasoning}},
	copyright = {Creative Commons Attribution Non Commercial No Derivatives 4.0 International},
	url = {https://arxiv.org/abs/2505.24478},
	doi = {10.48550/ARXIV.2505.24478},
	urldate = {2025-09-02},
	publisher = {arXiv},
	author = {Markovic, Vasilije and Obradovic, Lazar and Hajdu, Laszlo and Pavlovic, Jovan},
	year = {2025},
	note = {Version Number: 1},
}

@article{sequeda_knowledge_2025,
	title = {Knowledge {Graphs} as a source of trust for {LLM}-powered enterprise question answering},
	volume = {85},
	issn = {15708268},
	url = {https://linkinghub.elsevier.com/retrieve/pii/S1570826824000441},
	doi = {10.1016/j.websem.2024.100858},
	language = {en},
	urldate = {2025-09-02},
	journal = {Journal of Web Semantics},
	author = {Sequeda, Juan and Allemang, Dean and Jacob, Bryon},
	month = may,
	year = {2025},
	pages = {100858},
}

@misc{schilling-wilhelmi_text_2024,
	title = {From {Text} to {Insight}: {Large} {Language} {Models} for {Materials} {Science} {Data} {Extraction}},
	copyright = {Creative Commons Attribution 4.0 International},
	shorttitle = {From {Text} to {Insight}},
	url = {https://arxiv.org/abs/2407.16867},
	doi = {10.48550/ARXIV.2407.16867},
	urldate = {2025-09-02},
	publisher = {arXiv},
	author = {Schilling-Wilhelmi, Mara and Ríos-García, Martiño and Shabih, Sherjeel and Gil, María Victoria and Miret, Santiago and Koch, Christoph T. and Márquez, José A. and Jablonka, Kevin Maik},
	year = {2024},
	note = {Version Number: 2},
}

@article{zhang_exploring_2025,
	title = {Exploring the role of large language models in the scientific method: from hypothesis to discovery},
	volume = {1},
	issn = {3005-1460},
	shorttitle = {Exploring the role of large language models in the scientific method},
	url = {https://www.nature.com/articles/s44387-025-00019-5},
	doi = {10.1038/s44387-025-00019-5},
	language = {en},
	number = {1},
	urldate = {2025-09-02},
	journal = {npj Artificial Intelligence},
	author = {Zhang, Yanbo and Khan, Sumeer A. and Mahmud, Adnan and Yang, Huck and Lavin, Alexander and Levin, Michael and Frey, Jeremy and Dunnmon, Jared and Evans, James and Bundy, Alan and Dzeroski, Saso and Tegner, Jesper and Zenil, Hector},
	month = aug,
	year = {2025},
	pages = {14},
}

@misc{rueda_understanding_2025,
	title = {Understanding {LLM} {Scientific} {Reasoning} through {Promptings} and {Model}'s {Explanation} on the {Answers}},
	copyright = {Creative Commons Attribution 4.0 International},
	url = {https://arxiv.org/abs/2505.01482},
	doi = {10.48550/ARXIV.2505.01482},
	urldate = {2025-09-02},
	publisher = {arXiv},
	author = {Rueda, Alice and Hassan, Mohammed S. and Perivolaris, Argyrios and Teferra, Bazen G. and Samavi, Reza and Rambhatla, Sirisha and Wu, Yuqi and Zhang, Yanbo and Cao, Bo and Sharma, Divya and Krishnan, Sridhar and Bhat, Venkat},
	year = {2025},
	note = {Version Number: 2},
}

@misc{naveed_comprehensive_2023,
	title = {A {Comprehensive} {Overview} of {Large} {Language} {Models}},
	copyright = {Creative Commons Attribution 4.0 International},
	url = {https://arxiv.org/abs/2307.06435},
	doi = {10.48550/ARXIV.2307.06435},
	urldate = {2025-09-02},
	publisher = {arXiv},
	author = {Naveed, Humza and Khan, Asad Ullah and Qiu, Shi and Saqib, Muhammad and Anwar, Saeed and Usman, Muhammad and Akhtar, Naveed and Barnes, Nick and Mian, Ajmal},
	year = {2023},
	note = {Version Number: 10},
}

@misc{zhao_survey_2023,
	title = {A {Survey} of {Large} {Language} {Models}},
	copyright = {arXiv.org perpetual, non-exclusive license},
	url = {https://arxiv.org/abs/2303.18223},
	doi = {10.48550/ARXIV.2303.18223},
	urldate = {2025-09-02},
	publisher = {arXiv},
	author = {Zhao, Wayne Xin and Zhou, Kun and Li, Junyi and Tang, Tianyi and Wang, Xiaolei and Hou, Yupeng and Min, Yingqian and Zhang, Beichen and Zhang, Junjie and Dong, Zican and Du, Yifan and Yang, Chen and Chen, Yushuo and Chen, Zhipeng and Jiang, Jinhao and Ren, Ruiyang and Li, Yifan and Tang, Xinyu and Liu, Zikang and Liu, Peiyu and Nie, Jian-Yun and Wen, Ji-Rong},
	year = {2023},
	note = {Version Number: 16},
}

@article{teubner_welcome_2023,
	title = {Welcome to the {Era} of {ChatGPT} et al.: {The} {Prospects} of {Large} {Language} {Models}},
	volume = {65},
	issn = {2363-7005, 1867-0202},
	shorttitle = {Welcome to the {Era} of {ChatGPT} et al.},
	url = {https://link.springer.com/10.1007/s12599-023-00795-x},
	doi = {10.1007/s12599-023-00795-x},
	language = {en},
	number = {2},
	urldate = {2025-09-02},
	journal = {Business \& Information Systems Engineering},
	author = {Teubner, Timm and Flath, Christoph M. and Weinhardt, Christof and Van Der Aalst, Wil and Hinz, Oliver},
	month = apr,
	year = {2023},
	pages = {95--101},
}

@misc{touvron_llama_2023,
	title = {Llama 2: {Open} {Foundation} and {Fine}-{Tuned} {Chat} {Models}},
	copyright = {arXiv.org perpetual, non-exclusive license},
	shorttitle = {Llama 2},
	url = {https://arxiv.org/abs/2307.09288},
	doi = {10.48550/ARXIV.2307.09288},
	urldate = {2025-09-02},
	publisher = {arXiv},
	author = {Touvron, Hugo and Martin, Louis and Stone, Kevin and Albert, Peter and Almahairi, Amjad and Babaei, Yasmine and Bashlykov, Nikolay and Batra, Soumya and Bhargava, Prajjwal and Bhosale, Shruti and Bikel, Dan and Blecher, Lukas and Ferrer, Cristian Canton and Chen, Moya and Cucurull, Guillem and Esiobu, David and Fernandes, Jude and Fu, Jeremy and Fu, Wenyin and Fuller, Brian and Gao, Cynthia and Goswami, Vedanuj and Goyal, Naman and Hartshorn, Anthony and Hosseini, Saghar and Hou, Rui and Inan, Hakan and Kardas, Marcin and Kerkez, Viktor and Khabsa, Madian and Kloumann, Isabel and Korenev, Artem and Koura, Punit Singh and Lachaux, Marie-Anne and Lavril, Thibaut and Lee, Jenya and Liskovich, Diana and Lu, Yinghai and Mao, Yuning and Martinet, Xavier and Mihaylov, Todor and Mishra, Pushkar and Molybog, Igor and Nie, Yixin and Poulton, Andrew and Reizenstein, Jeremy and Rungta, Rashi and Saladi, Kalyan and Schelten, Alan and Silva, Ruan and Smith, Eric Michael and Subramanian, Ranjan and Tan, Xiaoqing Ellen and Tang, Binh and Taylor, Ross and Williams, Adina and Kuan, Jian Xiang and Xu, Puxin and Yan, Zheng and Zarov, Iliyan and Zhang, Yuchen and Fan, Angela and Kambadur, Melanie and Narang, Sharan and Rodriguez, Aurelien and Stojnic, Robert and Edunov, Sergey and Scialom, Thomas},
	year = {2023},
	note = {Version Number: 2},
}

@article{buehler_generative_2024,
	title = {Generative {Retrieval}-{Augmented} {Ontologic} {Graph} and {Multiagent} {Strategies} for {Interpretive} {Large} {Language} {Model}-{Based} {Materials} {Design}},
	volume = {4},
	copyright = {https://creativecommons.org/licenses/by-nc-nd/4.0/},
	issn = {2694-2488, 2694-2488},
	url = {https://pubs.acs.org/doi/10.1021/acsengineeringau.3c00058},
	doi = {10.1021/acsengineeringau.3c00058},
	language = {en},
	number = {2},
	urldate = {2025-09-02},
	journal = {ACS Engineering Au},
	author = {Buehler, Markus J.},
	month = apr,
	year = {2024},
	pages = {241--277},
}

@article{buehler_accelerating_2024,
	title = {Accelerating scientific discovery with generative knowledge extraction, graph-based representation, and multimodal intelligent graph reasoning},
	volume = {5},
	issn = {2632-2153},
	url = {https://iopscience.iop.org/article/10.1088/2632-2153/ad7228},
	doi = {10.1088/2632-2153/ad7228},
	number = {3},
	urldate = {2025-09-02},
	journal = {Machine Learning: Science and Technology},
	author = {Buehler, Markus J},
	month = sep,
	year = {2024},
	pages = {035083},
}

@misc{wei_emergent_2022,
	title = {Emergent {Abilities} of {Large} {Language} {Models}},
	url = {http://arxiv.org/abs/2206.07682},
	doi = {10.48550/arXiv.2206.07682},
	urldate = {2025-09-02},
	publisher = {arXiv},
	author = {Wei, Jason and Tay, Yi and Bommasani, Rishi and Raffel, Colin and Zoph, Barret and Borgeaud, Sebastian and Yogatama, Dani and Bosma, Maarten and Zhou, Denny and Metzler, Donald and Chi, Ed H. and Hashimoto, Tatsunori and Vinyals, Oriol and Liang, Percy and Dean, Jeff and Fedus, William},
	month = oct,
	year = {2022},
	note = {arXiv:2206.07682 [cs]},
}

@misc{luu_generative_2025,
	title = {Generative {Artificial} {Intelligence} {Extracts} {Structure}-{Function} {Relationships} from {Plants} for {New} {Materials}},
	copyright = {Creative Commons Attribution Non Commercial No Derivatives 4.0 International},
	url = {https://arxiv.org/abs/2508.06591},
	doi = {10.48550/ARXIV.2508.06591},
	urldate = {2025-09-02},
	publisher = {arXiv},
	author = {Luu, Rachel K. and Deng, Jingyu and Ibrahim, Mohammed Shahrudin and Cho, Nam-Joon and Dao, Ming and Suresh, Subra and Buehler, Markus J.},
	year = {2025},
	note = {Version Number: 1},
}

@article{luu_bioinspiredllm_2024,
	title = {{BioinspiredLLM}: {Conversational} {Large} {Language} {Model} for the {Mechanics} of {Biological} and {Bio}‐{Inspired} {Materials}},
	volume = {11},
	issn = {2198-3844, 2198-3844},
	shorttitle = {{BioinspiredLLM}},
	url = {https://onlinelibrary.wiley.com/doi/10.1002/advs.202306724},
	doi = {10.1002/advs.202306724},
	language = {en},
	number = {10},
	urldate = {2025-09-02},
	journal = {Advanced Science},
	author = {Luu, Rachel K. and Buehler, Markus J.},
	month = mar,
	year = {2024},
	pages = {2306724},
}

@article{ghafarollahi_protagents_2024,
	title = {{ProtAgents}: protein discovery \textit{via} large language model multi-agent collaborations combining physics and machine learning},
	volume = {3},
	issn = {2635-098X},
	shorttitle = {{ProtAgents}},
	url = {https://xlink.rsc.org/?DOI=D4DD00013G},
	doi = {10.1039/D4DD00013G},
	language = {en},
	number = {7},
	urldate = {2025-09-02},
	journal = {Digital Discovery},
	author = {Ghafarollahi, Alireza and Buehler, Markus J.},
	year = {2024},
	pages = {1389--1409},
}

@article{ni_mechagents_2024,
	title = {{MechAgents}: {Large} language model multi-agent collaborations can solve mechanics problems, generate new data, and integrate knowledge},
	volume = {67},
	issn = {23524316},
	shorttitle = {{MechAgents}},
	url = {https://linkinghub.elsevier.com/retrieve/pii/S2352431624000117},
	doi = {10.1016/j.eml.2024.102131},
	language = {en},
	urldate = {2025-09-02},
	journal = {Extreme Mechanics Letters},
	author = {Ni, Bo and Buehler, Markus J.},
	month = mar,
	year = {2024},
	pages = {102131},
}

@article{cousins_high_2020,
	title = {The high persistence of {PFAS} is sufficient for their management as a chemical class},
	volume = {22},
	issn = {2050-7895},
	doi = {10.1039/d0em00355g},
	language = {eng},
	number = {12},
	journal = {Environmental Science. Processes \& Impacts},
	author = {Cousins, Ian T. and DeWitt, Jamie C. and Glüge, Juliane and Goldenman, Gretta and Herzke, Dorte and Lohmann, Rainer and Ng, Carla A. and Scheringer, Martin and Wang, Zhanyun},
	month = dec,
	year = {2020},
	pages = {2307--2312},
}

@article{shabanian_durable_2023,
	title = {Durable water- and oil-repellent textiles without long- or short-chain perfluoroalkylated substances},
	volume = {34},
	issn = {24685194},
	url = {https://linkinghub.elsevier.com/retrieve/pii/S2468519423004135},
	doi = {10.1016/j.mtchem.2023.101786},
	language = {en},
	urldate = {2025-09-02},
	journal = {Materials Today Chemistry},
	author = {Shabanian, Sadaf and Lahiri, Sudip Kumar and Soltani, Mohammad and Golovin, Kevin},
	month = dec,
	year = {2023},
	pages = {101786},
}

@article{leung_physicochemical_2023,
	title = {Physicochemical properties and interactions of perfluoroalkyl substances ({PFAS}) - {Challenges} and opportunities in sensing and remediation},
	volume = {905},
	issn = {00489697},
	url = {https://linkinghub.elsevier.com/retrieve/pii/S0048969723053895},
	doi = {10.1016/j.scitotenv.2023.166764},
	language = {en},
	urldate = {2025-09-02},
	journal = {Science of The Total Environment},
	author = {Leung, Shui Cheung Edgar and Wanninayake, Dushanthi and Chen, Dechao and Nguyen, Nam-Trung and Li, Qin},
	month = dec,
	year = {2023},
	pages = {166764},
}

@article{zheng_per-_2021,
	title = {Per- and {Polyfluoroalkyl} {Substances} ({PFAS}) in {Breast} {Milk}: {Concerning} {Trends} for {Current}-{Use} {PFAS}},
	volume = {55},
	copyright = {https://doi.org/10.15223/policy-029},
	issn = {0013-936X, 1520-5851},
	shorttitle = {Per- and {Polyfluoroalkyl} {Substances} ({PFAS}) in {Breast} {Milk}},
	url = {https://pubs.acs.org/doi/10.1021/acs.est.0c06978},
	doi = {10.1021/acs.est.0c06978},
	language = {en},
	number = {11},
	urldate = {2025-09-02},
	journal = {Environmental Science \& Technology},
	author = {Zheng, Guomao and Schreder, Erika and Dempsey, Jennifer C. and Uding, Nancy and Chu, Valerie and Andres, Gabriel and Sathyanarayana, Sheela and Salamova, Amina},
	month = jun,
	year = {2021},
	pages = {7510--7520},
}

@article{henderson_analysis_2024,
	title = {Analysis of {Legacy} and {Novel} {Neutral} {Per}- and {Polyfluoroalkyl} {Substances} in {Soils} from an {Industrial} {Manufacturing} {Facility}},
	volume = {58},
	copyright = {https://doi.org/10.15223/policy-001},
	issn = {0013-936X, 1520-5851},
	url = {https://pubs.acs.org/doi/10.1021/acs.est.3c10268},
	doi = {10.1021/acs.est.3c10268},
	language = {en},
	number = {24},
	urldate = {2025-09-02},
	journal = {Environmental Science \& Technology},
	author = {Henderson, W. Matthew and Evich, Marina G. and Washington, John W. and Ward, Thomas T. and Schumacher, Brian A. and Zimmerman, John H. and Kim, Yung D. and Weber, Eric J. and Williams, Alan C. and Smeltz, Marci G. and Glinski, Donna A.},
	month = jun,
	year = {2024},
	pages = {10729--10739},
}

\end{document}